\documentclass[conference]{IEEEtran}
\IEEEoverridecommandlockouts
\usepackage{cite}
\usepackage{amsmath,amssymb,amsfonts}
\usepackage{algorithmic}
\usepackage{graphicx}
\usepackage{textcomp}
\usepackage{xcolor}

\usepackage{color}
\usepackage{subfigure}
\usepackage{float}
\usepackage{multirow}
\usepackage{hyperref}
\hypersetup{
    colorlinks=true,
    linkcolor=blue,
    filecolor=magenta,      
    urlcolor=cyan,
    pdftitle={Overleaf Example},
    pdfpagemode=FullScreen,
    }

\def\BibTeX{{\rm B\kern-.05em{\sc i\kern-.025em b}\kern-.08em
    T\kern-.1667em\lower.7ex\hbox{E}\kern-.125emX}}
\begin{document}

% \title{Conference Paper Title*\\
% {\footnotesize \textsuperscript{*}Note: Sub-titles are not captured in Xplore and
% should not be used}
% \thanks{Identify applicable funding agency here. If none, delete this.}
% }

\title{PointNorm: Dual Normalization is All You Need for Point Cloud Analysis\\}

% \author{\IEEEauthorblockN{Anonymous Authors}}

% TODO: modify the authors below 

\author{\IEEEauthorblockN{Shen Zheng}
\IEEEauthorblockA{\textit{School of Computer Science} \\
\textit{Carnegie Mellon University}\\
Pittsburgh, USA \\
shenzhen@andrew.cmu.edu}
\and
\IEEEauthorblockN{Jinqian Pan$^{*}$}
\IEEEauthorblockA{\textit{Center for Data Science} \\
\textit{New York University}\\
New York, USA \\
jp6218@nyu.edu}
\and
\IEEEauthorblockN{Changjie Lu$^{*}$, Gaurav Gupta}
\IEEEauthorblockA{\textit{College of Science and Technology} \\
\textit{Wenzhou-Kean University}\\
Wenzhou, China \\
\{lucha, ggupta\}@kean.edu}
\thanks{$^{*}$ indicates equal contribution}
}

% \author{\IEEEauthorblockN{Shen Zheng}
% \IEEEauthorblockA{\textit{School of Computer Science} \\
% \textit{Carnegie Mellon University}\\
% Pittsburgh, USA \\
% shenzhen@andrew.cmu.edu}
% \and
% \IEEEauthorblockN{Jinqian Pan}
% \IEEEauthorblockA{\textit{Center for Data Science} \\
% \textit{New York University}\\
% New York, USA \\
% jp6218@nyu.edu}
% \and
% \IEEEauthorblockN{Changjie Lu}
% \IEEEauthorblockA{\textit{College of Science and Technology} \\
% \textit{Wenzhou-Kean University}\\
% Wenzhou, China \\
% lucha@kean.edu}
% \and
% \IEEEauthorblockN{Gaurav Gupta}
% \IEEEauthorblockA{\textit{College of Science and Technology} \\
% \textit{Wenzhou-Kean University}\\
% Wenzhou, China \\
% ggupta@kean.edu}
% }

\maketitle

\begin{abstract}
Point cloud analysis is challenging due to the irregularity of the point cloud data structure. Existing works typically employ the ad-hoc sampling-grouping operation of PointNet++, followed by sophisticated local and/or global feature extractors for leveraging the 3D geometry of the point cloud. Unfortunately, the sampling-grouping operations do not address the point cloud's irregularity, whereas the intricate local and/or global feature extractors led to poor computational efficiency. In this paper, we introduce a novel DualNorm module after the sampling-grouping operation to effectively and efficiently address the irregularity issue. The DualNorm module consists of Point Normalization, which normalizes the grouped points to the sampled points, and Reverse Point Normalization, which normalizes the sampled points to the grouped points. The proposed framework, PointNorm, utilizes local mean and global standard deviation to benefit from both local and global features while maintaining a faithful inference speed. Experiments show that we achieved excellent accuracy and efficiency on ModelNet40 classification, ScanObjectNN classification, ShapeNetPart Part Segmentation, and S3DIS Semantic Segmentation. Code is available at \href{https://github.com/ShenZheng2000/PointNorm-for-Point-Cloud-Analysis}{https://github.com/ShenZheng2000/PointNorm-for-Point-Cloud-Analysis}.
\end{abstract}


\section{Introduction}

% \textit{``Simplicity is the ultimate sophistication.'' -- Leonardo da Vinci.}

% Point Cloud is popular because of its wide application  
% P.C. is difficult because its is irregular 
A point cloud is a group of points for describing the object shapes in the 3D space. Due to its importance for many downstream applications, such as autonomous driving \cite{chen20203d}, virtual and augmented reality \cite{ni2018point}, and robotics \cite{pomerleau2015review}, point cloud analysis has recently gained enormous popularity in the computer vision community. However, point clouds are irregular (i.e., unevenly distributed), which makes learning 3D geometry challenging \cite{qi2017pointnet++}.

% (i.e., have different densities across different regions)

% PointNet/ PointNet++ -> Local and Global Approachs and their limitations
The pioneering work PointNet \cite{qi2017pointnet} attends each point in the point cloud independently with Multi-Layer Perceptions (MLPs). The follow-up work PointNet++ \cite{qi2017pointnet++} introduces set abstraction layers to exploit the local geometry. Specifically, the set abstraction layers consist of a sampling layer, a grouping layer, and a PointNet layer for non-linear mapping. The sampling layer (e.g., Farthest Point Sampling (FPS) \cite{moenning2003fast}) is used to select the representative points, whereas the grouping layer (e.g., K-Nearest Neighbors (KNN) \cite{peterson2009k}) is employed to select the points closest to the representative points.

% Head Figure for Explanation
\begin{figure}[t]
    \centering
    \includegraphics[width=8.5cm]{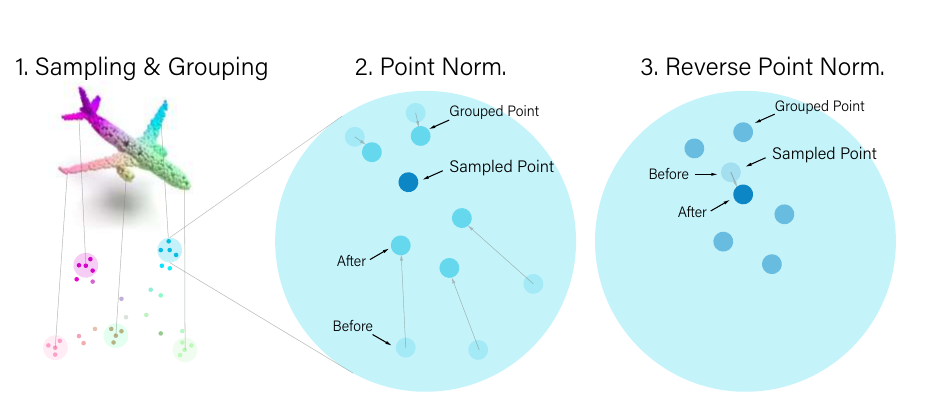}
    \caption{Overview of DualNorm (Point Normalization and Reverse Point Normalization) in our PointNorm framework. After sampling and grouping, we first normalize the grouped points to the sampled points and then normalize the sampled points to the grouped point. In this way, we address the irregularity of the point cloud and facilitate the learning for subsquent layers.}
    \label{fig:Model_Head}
\end{figure}

% current progresses
With the success of PointNet++, a handful of works have attempted to exploit the local geometry of point cloud by incorporating convolution \cite{wu2019pointconv, xu2021paconv}, graph \cite{wang2019dynamic, lin2020convolution}, transformers \cite{guo2021pct, zhao2021point} or geometric methods \cite{ma2022rethinking, ran2022surface} in the non-linear mapping layers. Unlike the local geometry, the global point features bring long-range semantics, which benefit part segmentation and semantic segmentation \cite{yan2020pointasnl, xiang2021walk, bello2020deep}. The global geometry exploration of point cloud often involves adaptive sampling \cite{yan2020pointasnl}, mesh \cite{lahav2020meshwalker}, curve \cite{xiang2021walk}, or surfaces \cite{ran2022surface}. 

% limitations of sampling-grouping operations
Despite their improvements in classification and segmentation accuracy, both approaches follow PointNet++'s sampling-grouping operations, which fail to represent the 3D geometry of the irregular point cloud. Without addressing the irregularity issue, these approaches engage in a competition for complex feature extractors and optimization procedures, resulting in extended inference latency. This is unacceptable for large-scale point cloud analysis \cite{landrieu2018large} and mobile point cloud analysis \cite{pomerleau2015review}, which both emphasize computational efficiency.

\begin{figure*}[t]
    \centering
    \includegraphics[width=19cm]{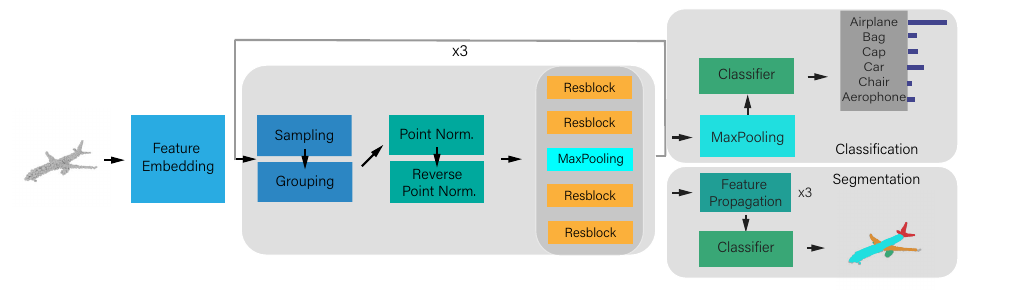}
    \caption{PointNorm for shape classification and part segmentation. Given an input point cloud, PointNorm first embeds the point features and then utilizes sampling-grouping and DualNorm (Point Normalization and Reverse Point Normalization) to normalize sampled and grouped points. The Residual Blocks exploits the hierarchical features to assist the classification and the segmentation head to model the correct category.}
    \label{fig:Model_WorkFlow}
\end{figure*}

In this paper, we seek to eliminate complicated local and/or global feature extractors and develop a simple, intuitive network for point cloud analysis. Inspired by how normalization serves as a pre-processing step to address irregular data in 2D image processing \cite{patro2015normalization, ulyanov2016instance}, we propose to add normalization \textit{after} sampling-grouping, but \textit{before} the non-linear mapping layers. In this way, we can enjoy the efficiency of the sampling-grouping operation while preventing irregular data from passing into the subsequent non-linear mapping layers. Since sampled and grouped points are both crucial for point cloud feature learning, it is hard to decide which one to normalize. Instead of discarding any one of them, we attempt to utilize both points using a novel module called DualNorm. As shown in Fig. \ref{fig:Model_Head}, the DualNorm module first normalizes the grouped points to the sampled points and then normalizes the sampled points to the grouped points. Using local mean and global standard deviation in the normalization allows the proposed method to leverage local and global features while enjoying a faithful inference speed. The contributions of this paper can be summarized as below:

\begin{itemize}[leftmargin=1em,topsep=0pt,partopsep=0pt]
     \item We propose a new framework named PointNorm. With local mean and global standard deviation, PointNorm eliminates the need for sophisticated local and global feature extractors, thereby significantly enhancing the computational efficiency. 
    \item We introduce a novel, plug-and-play style module called DualNorm. By normalizing sampled points and grouped points to each other, the DualNorm module uses an adaptive push-and-pull strategy to optimize the point cloud density, thereby addressing the point cloud irregularity. 
    \item We provide analysis to show that DualNorm can remarkably improve the loss stability and gradient stability with a trivial increase in computational complexity.
    
    % By normalizing sampled points and grouped points to each other, we address the irregularity of the point cloud and make learning point cloud features easier for the subsequent layers. 
    
    % \item We design a plug-and-play style Controllable Residual Block (C-Resblock) to effectively trade-off between model performance and model size.
    \item We demonstrate that PointNorm achieves compelling performances on point cloud classification and segmentation.
\end{itemize}

The rest of the paper is organized as follows. Section \ref{sec:rel_work} provides a comprehensive review of related works. Section \ref{sec:pro_method} presents the key components of the proposed methods. Section \ref{sec:experiments} outlines the experimental details and provides qualitative and quantitative comparisons. Section \ref{sec:conclusion} summarizes the findings and discusses the future works.

% \footnote{We do not consider normalization before the sampling-grouping operation as it is highly time-consuming to process all point cloud at the beginning.}

% Finally, we designed a novel Controllable Residual \cite{he2016deep} Block (C-Resblock). Inspired by squeeze-excitation network \cite{hu2018squeeze} and inverted residual block \cite{sandler2018mobilenetv2}, C-Resblock has a bottleneck ratio that can either squeeze or expand the model without significantly affecting the model performance. This allow us to develop lightweight versions with an elegant trade-off between model performance and model size. 

% TODO: summary rest of the articles 

% % NOTE: merge this figure with the figure above !!!
% \begin{figure*}[t]
%     \centering
%     \includegraphics[width=17.5cm]{Model_Figure/Model_WorkFlow_Seg.pdf}
%     \caption{The workflow of PointNorm for part segmentation tasks. Compared with PointNorm for classification, We add three feature propagation layers \cite{qi2017pointnet++} before the classifier to upsample the point features, and we utilize a softmax function after the classifier to calculate the scores for the negative log-likelihood loss.}
%     \label{fig:Model_WorkFlow_Seg}
% \end{figure*}

\section{Related Work} \label{sec:rel_work}
    \subsection{Point Cloud Analysis}
    % voxel, multi-view, and their limitations -> directly on points 
    Due to the irregularity of the point cloud data structure, prior works attempt to transform the raw point cloud into intermediate voxels \cite{maturana2015voxnet, zhou2018voxelnet}, or multi-view images \cite{you2018pvnet, hamdi2021mvtn}, thereby converting the 3D challenge to a well-investigated 2D task. However, these transformations are accompanied by heavy computation and loss of shape details \cite{yang2019std, bello2020deep}. PointNet \cite{qi2017pointnet} is the pioneering method that operates directly upon the raw point cloud. Based upon PointNet, PointNet++ \cite{qi2017pointnet++} add a sampling-grouping operation before the MLP layers to extract the local geometry. Like PointNet++, and the follow-up works \cite{li2018pointcnn, liu2019relation, wu2019pointconv, ma2022rethinking}, our PointNorm utilizes the sampling-grouping operation with a MLP-based architecture. Unlike previous works, PointNorm leverages normalization between the sampling-grouping layers and the MLP layers to address the irregular point cloud.
    
    \subsection{Point Cloud Geometry Exploration}
    Inspired by the success of PointNet++, recent research has greatly explored the point cloud local geometry , using convolutions \cite{wu2019pointconv, xu2021paconv}, graphs \cite{wang2019dynamic, lin2020convolution}, transformers \cite{guo2021pct, lin2020convolution}. The most prominent convolution-based method is PointConv \cite{wu2019pointconv}, which utilizes MLPs and density functions to obtain the weight function for a given point. Unlike convolution-based methods, graph-based methods (e.g., EdgeConv \cite{wang2019dynamic}) treat each point as a graph node and utilize graph edges to capture the local geometry between a specific point and its neighbors, whereas the transformer-based methods like Point Transformer \cite{zhao2021point} utilize self-attention to extract the local neighborhoods around each point. 
    
    % In comparison, the geometry-based is more customized. For example, PointMLP \cite{ma2022rethinking} utilizes a geometric affine module to adaptively translate the points in a local region, whereas Repsurf \cite{ran2022surface} leverages representative surfaces to describe the points in a local region.
    
    The attempt to incorporate global geometry in point cloud analysis is predominantly inspired by the success of non-local network \cite{wang2018non} in 2D image processing. The main challenge of point cloud global geometry extraction is how to effectively aggregate the local and the global features \cite{xiang2021walk, guo2020deep}. To this end, PointASNL \cite{yan2020pointasnl} presents a local-nonlocal module with adaptive sampling to capture the local and global dependencies of the sampled point. MeshWalker \cite{lahav2020meshwalker} explores local and global geometry through random walks along the mesh surfaces. CurveNet \cite{xiang2021walk} develops a curve-based guided walk strategy to group and aggregate local-nonlocal point features. 
    
    % The local and global methods mentioned above require sophisticated searching and optimization to extract and aggregate the local and global features, leading to poor inference latency. 
    The local and global approaches mentioned above require advanced searching and optimization to collect and combine features, causing poor inference latency. In comparison, the proposed PointNorm enjoys excellent computational efficiency because it only needs normalization operations using the mean of the local points and the standard deviation of the global points (i.e., all points).

\section{Proposed Methods} \label{sec:pro_method}
In this section, we first introduce DualNorm and then employ standard deviation analysis and optimization landscape analysis to show the superiority of DualNorm. After that, we illustrate the model architecture for PointNorm. Finally, we introduce a lightweight version named PointNorm-Tiny.

\subsection{DualNorm} \label{DualNorm}

\subsubsection{Preliminaries} 
PointNet++ \cite{qi2017pointnet++} performs both sampling and grouping operations on the irregular point cloud. Consequently, the sampled and grouped points are both irregular. Our DualNorm modules consist of Point Normalization (PN) and Reverse Point Normalization (RPN). Specifically, PN aims to address the irregularity of the grouped points, whereas RPN aims to address the irregularity of the sampled points. 

\subsubsection{Notations}
Suppose $x_{s}$ represents the sampled points; $x_{g}$ represents the grouped points; $k$ is the number of neighbors in the KNN algorithm \cite{peterson2009k}; $n$ is the number of points in a point cloud; $d$ is the dimension (i.e., the channel number at a specific layer); $\alpha_{1}$ and $\alpha_{2}$ are learnable parameters initialized with a value of 1; $\beta_{1}$ and $\beta_{2}$ are learnable parameters initialized with a value of 0; $\varepsilon$ is set as $10^{-5}$ to promote numerical stability (e.g., avoid division by zero).

% highlight local and global choices (different)

% highlight order (PN -> RPN)

\subsubsection{Point Normalization} PN normalizes $x_{g}$ to $x_{s}$. The mean $\mu_{s}$ and the standard deviation $\sigma_{1}$ for PN are:

% The local mean is $\mu_{s}=x_{s}$, and the global mean is $\mu_{s}=\frac{1}{n d} \sum_{b=1}^{n} \sum_{c=1}^{d} x_{s}$. 

\begin{equation}
     \mu_{1}= 
        \begin{cases}
            x_{s} & $local$ \\
            \frac{1}{n d} \sum_{b=1}^{n} \sum_{c=1}^{d} x_{s} & $global$
        \end{cases}
\end{equation}

% The local standard deviation is $\sigma_{1}=\sqrt{\frac{1}{k} \sum_{a=1}^{k}\left[x_{g_{a}}-\mu_{s}\right]^{2}}$, and the global standard deviation is $\sigma_{1}=\sqrt{\frac{1}{k n d} \sum_{a=1}^{k} \sum_{b=1}^{n}  \sum_{c=1}^{d}\left[x_{g_{a,b,c}}-\mu_{s}\right]^{2}}$. 

\begin{equation}
    \sigma_{1}= 
        \begin{cases}
            \sqrt{\frac{1}{k} \sum_{a=1}^{k}\left[x_{g_{a}}-\mu_{1}\right]^{2}} & $local$ \\
            \sqrt{\frac{1}{k n d} \sum_{a=1}^{k} \sum_{b=1}^{n}  \sum_{c=1}^{d}\left[x_{g_{a,b,c}}-\mu_{1}\right]^{2}} & $global$
        \end{cases}
\end{equation}

The scale-and-shift strategy in batch normalization \cite{ioffe2015batch} is employed to promote non-linearity and to make the PN layers trainable in the backpropagation process. The scale-and-shift normalization for PN is 

\begin{equation}
    x_{g} \leftarrow \alpha_{1} * \frac{x_{g}-\mu_{1}}{\sigma_{1}+\varepsilon}+\beta_{1}
\end{equation}

\subsubsection{Reverse Point Normalization} RPN normalizes $x_{s}$ to $x_{g}$. The mean $\mu_{g}$ and the standard deviation\footnote{We are unable to calculate the local standard deviation with one sampled point in a KNN-group. Instead, we consider the L1 distance between $\mu_{g}$ and $x_{s}$, and use square root to promote non-linearity.} $\sigma_{2}$ for RPN are:

% The local mean is $\mu_{g}=\frac{1}{k} \sum_{a=1}^{k} x_{g}$, and the global mean is $\mu_{g}=\frac{1}{k n d} \sum_{a=1}^{k} \sum_{b=1}^{n} \sum_{c=1}^{d} x_{g}$. 

\begin{equation}
     \mu_{2}= 
        \begin{cases}
            \frac{1}{k} \sum_{a=1}^{k} x_{g} & $local$ \\
            \frac{1}{k n d} \sum_{a=1}^{k} \sum_{b=1}^{n} \sum_{c=1}^{d} x_{g} & $global$
        \end{cases}
\end{equation}

\begin{equation}
    \sigma_{2}= 
        \begin{cases}
             \sqrt{\left|x_{s}-\mu_{2}\right|+\varepsilon} & $local$ \\
             \sqrt{\frac{1}{k n d} \sum_{a=1}^{k} \sum_{b=1}^{n}  \sum_{c=1}^{d}\left[x_{s_{a,b,c}}-\mu_{2}\right]^{2}} & $global$
        \end{cases}
\end{equation}

%  The local standard deviation is therefore formulated as $\sigma_{2}=\sqrt{\left|x_{s}-\mu_{g}\right|+\varepsilon}$. Since there is always more than one sampled point in a batch, the global standard deviation can be formulated naturally as we did for PN. The global standard deviation is $\sigma_{2}=\sqrt{\frac{1}{k n d} \sum_{a=1}^{k} \sum_{b=1}^{n}  \sum_{c=1}^{d}\left[x_{s_{a,b,c}}-\mu_{g}\right]^{2}}$
 
The scale-and-shift normalization for RPN is

\begin{equation}
    x_{s} \leftarrow \alpha_{2} * \frac{x_{s}-\mu_{2}}{\sigma_{2}+\varepsilon}+\beta_{2}
\end{equation}

% according to the local/global standard deviation $\sigma_i (i=1,2)$ and the learnable parameter $\alpha_i (i=1,2)$.

% Let's take PN as an example since RPN can be done in a similar way due to the symmetry of PN and RPN. 

% Model_Alpha
\begin{figure}[t]
    \centering
    \includegraphics[width=8.5cm]{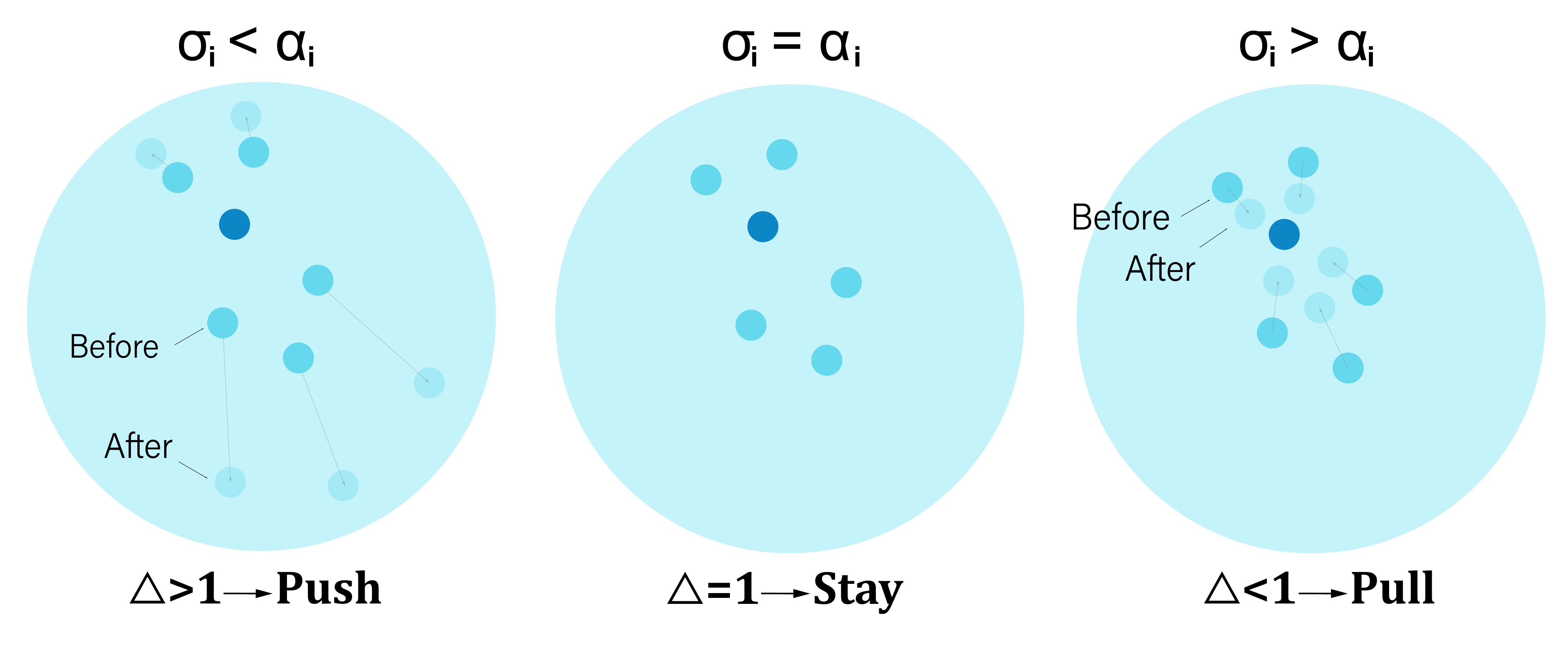}
    \caption{PointNorm's `push-and-pull' strategy for optimizing the point cloud density. When $\Delta > 1$ (left column), PointNorm pushes points apart, which increases the standard deviation and reduces the point cloud density; when $\Delta < 1$ (right column), PointNorm pulls points together, which reduces the standard deviation and increase the point cloud density; when $\Delta = 1$ (middle column), no action is performed, and both the standard deviation and the point cloud density stay the same. }
    \label{fig:Model_Alpha}.
\end{figure}

\subsubsection{Standard Deviation Analysis}
We analyze the standard deviation change of point cloud to understand how DualNorm address the irregularity of point cloud. Suppose that the points before and after DualNorm is $x$ and $\hat{x}$, and the standard deviation for $x$ and $\hat{x}$ is $\sigma_{x}$ and $\sigma_{\hat{x}}$. According to the scale-and-shift normalization, we have:
\begin{equation} \label{eq:1}
    \hat{x} = \alpha_{i} * \frac{x-\mu_{i}}{\sigma_{i}+\varepsilon}+\beta_{i}
\end{equation}

If we ignore the tiny $\varepsilon$, we can express $\sigma_{\hat{x}}$ as:
\begin{equation} \label{eq:2}
    \sigma_{\hat{x}} = \frac{\alpha_{i}}{\sigma_{i}} * \sigma_{x}
\end{equation}

Suppose that the standard deviation change ratio is defined as $\Delta = \frac{\sigma_{\hat{x}}}{\sigma_{x}}$. Equ. \ref{eq:2} can therefore be written as:
\begin{equation} \label{eq:3}
    \Delta = \frac{\alpha_{i}}{\sigma_{i}}
\end{equation}

% TODO: polish writing here
Since $\alpha_i$ is a learnable parameter, we are able to adopt customized normalization strategies for three scenarios (See Fig. \ref{fig:Model_Alpha}) to address the irregularity of the point cloud.

\begin{figure}[t]
    \centering

     \includegraphics[width=4.1cm]{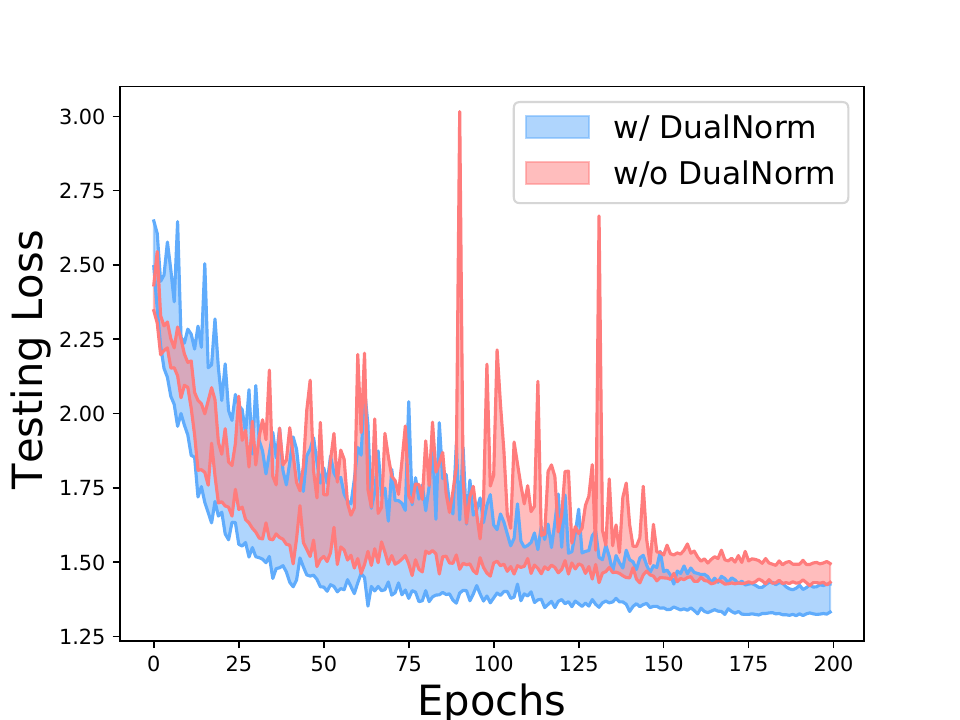}
     \includegraphics[width=4.1cm]{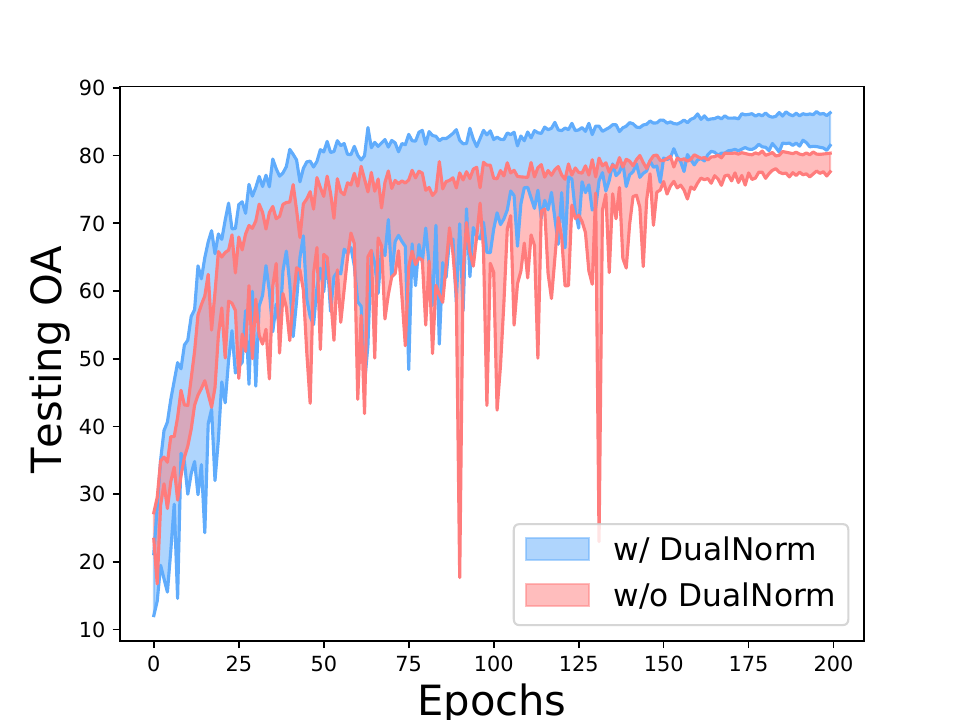}
     \includegraphics[width=4.1cm]{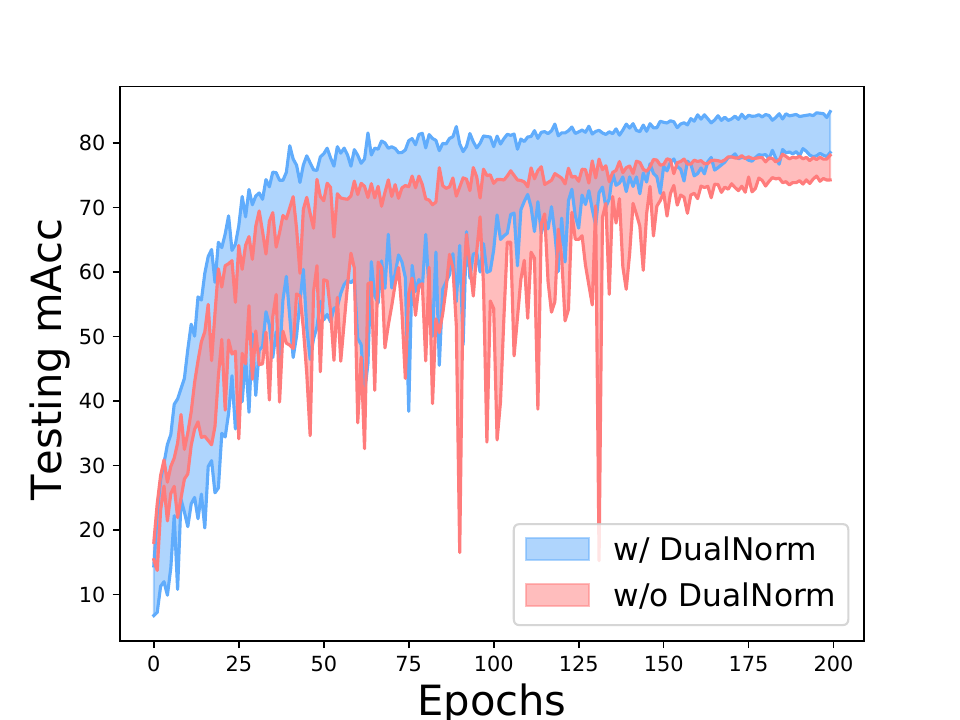}
     \includegraphics[width=4.1cm]{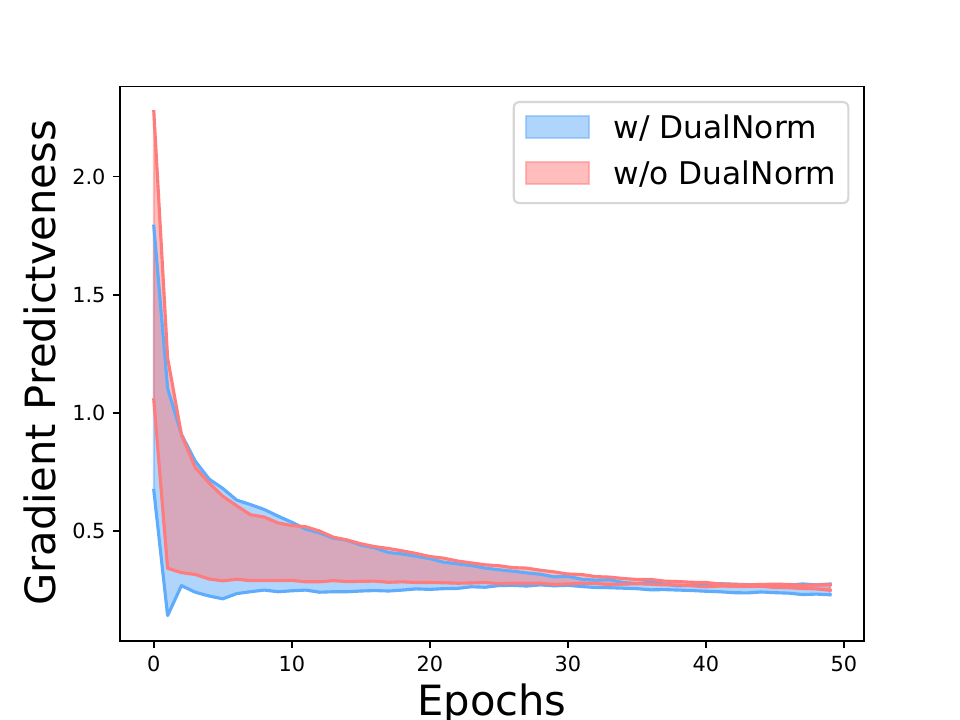}

    \caption{Optimization Landscape Analysis with Testing Loss$\downarrow$, OA$\uparrow$, mAcc$\uparrow$, and Gradient Predictiveness$\downarrow$. Zoom in for a better view.}
     \label{fig:Opt_Land}
\end{figure}

\subsubsection{Optimization Landscape Analysis}

% Explain Metric Setups
    % common: different learning rate (how much)
        % min -> lower bound
        % max -> upper bound
        % between -> bounded area
    % Testing loss (and OA and mAcc)
        % examine loss stability (Lipschitzness)
    % l2 difference between sub. gradients
        % examine gradient stability (predictiveness)
% Describe Figures
% illustrate implications

% Declare Theorems

We perform an optimization landscape analysis \cite{santurkar2018does} on ScanObjectNN \cite{uy2019revisiting} to examine the effectiveness of DualNorm. Two PointNorm variants, one with DualNorm, and the other without DualNorm, are investigated with different initial learning rates (0.005, 0.01, 0.002). For each variant, the learning rate configuration leading to the lowest/highest score of a metric is plotted as the lower/upper bound for the shaded region in Fig. \ref{fig:Opt_Land}.

For evaluation metrics, we choose testing loss, OA and mAcc to examine the loss stability (Lipschitzness) and gradient predictiveness (the L2 gradient change) to examine the gradient stability. Fig. \ref{fig:Opt_Land} shows that PointNorm with DualNorm has less fluctuation and better scores for loss, OA, and mAcc, which means DualNorm help reduce the Lipschitz constant and improve the loss stability during training. Besides, PointNorm with DualNorm has lower gradient predictiveness, especially during the early-stage training. This result indicates DualNorm help reduce the L2 gradient change and ensures better gradient stability during training.

\subsection{PointNorm} We show the workflow of PointNorm in Fig. \ref{fig:Model_WorkFlow}. PointNorm consists of five steps: feature embedding, sampling-grouping, DualNorm (PN+RPN), non-linear mapping, and classification/segmentation. The feature embedding step aims to raise the channel number so that the network contains more learnable parameters. Inspired by PointNet++ \cite{qi2017pointnet++}, we repeat the sampling-grouping, the DualNorm, and the non-linear mapping step three times to extract the hierarchical features with different receptive fields. The FPS \cite{moenning2003fast} is used for point sampling, whereas KNN with k = 24 \cite{peterson2009k} is used for point grouping. Following sampling-grouping, we leverage DualNorm to normalize the sampled and the grouped points. Following DualNorm, the sampled and the grouped points are concatenated before entering the subsequent non-linear mapping step. The non-linear mapping step consists of four Residual Blocks and one max pooling layer in the middle for channel number reduction. Finally, inspired by previous works \cite{qi2017pointnet, qi2017pointnet++, zhou2018voxelnet, wang2019dynamic}, we use max pooling to preserve the salient features for classification, and progressive feature propagation to upsample the point features for segmentation.

% FPS: qi2017pointnet++, wang2018local, ma2022rethinking
% KNN: li2018pointcnn, wang2018local, wang2019dynamic, ma2022rethinking

% The supplementary material will include PointNorm's architecture for part segmentation tasks.

% \subsection{C-ResBlock}

% Ablation Result for PointNorm at ScanObjectNN Dataset.
% \cline{2-8}

\subsection{PointNorm-Tiny} \label{PointNorm-Tiny}
 PointNorm has excellent inference latency and classification accuracy, but many amount of parameters (12.63M). This motivates us to design a lightweight version suitable for mobile deployment. The lightweight version should have few parameters, a high inference latency, and a good classification accuracy. To create PointNorm-Tiny, we make three adjustments. First, we reduce the bottleneck ratio from 1.00 to 0.25. Second, we lower the feature embedding dimension from 64 to 32. Third, we reduce ResBlock by 50\%. PointNorm-Tiny has only 0.68M parameters, infers faster than PointNet++, and delivers delivers faithful classification accuracy in the experiments.

% Indeed, with the proposed C-ResBlock, we can easily squeeze PointNorm by decreasing the bottleneck ratio. As shown in Table \ref{tab:Ablation_ScanObjectNN}, PointNorm with a bottleneck ratio of 0.25 can reduce the number of parameters to 4.65M, which is much smaller than 12.63M. However, we are ambitious to design the \textit{smallest} and the \textit{faster} model which is essential for mobile deployment. Therefore, apart from 

% motivation: good inference speed but relatively large model size
% how to adapt for mobile deployment?

% designs
    % bottleneck ratio (1->0.25)
    % block numbers (half)
    % embed dim (64->32)

\section{Experiments} \label{sec:experiments}

% % Acc (mAcc, OA) Curve for Ablation Study
%     \begin{figure}
%         \centering
        
%         % for Layer
%         \includegraphics[width=4.1cm]{Ablation/AccCurve/Ablation_Layer_OA.pdf}
%         \includegraphics[width=4.1cm]{Ablation/AccCurve/Ablation_Layer_mAcc.pdf}
        
%         % for Block (skip for now)
%         % \includegraphics[width=4.3cm]{Ablation/AccCurve/Ablation_Block_OA.pdf}
%         % \includegraphics[width=4.3cm]{Ablation/AccCurve/Ablation_Block_mAcc.pdf}
        
%         % for Local/Global
%         \includegraphics[width=4.1cm]{Ablation/AccCurve/Ablation_Local_Global_OA.pdf}
%         \includegraphics[width=4.1cm]{Ablation/AccCurve/Ablation_Local_Global_mAcc.pdf}
        
%         % for Norm
%         \includegraphics[width=4.1cm]{Ablation/AccCurve/Ablation_Norm_OA.pdf}
%         \includegraphics[width=4.1cm]{Ablation/AccCurve/Ablation_Norm_mAcc.pdf}

%         \caption{Testing Accuracy Curve for different variants of PointNorm. Left/Right Column: Overall Accuracy (OA)/Mean Accuracy (mAcc) to Training Epochs. Both OA and mAcc are reported in \%. Zoom in for a better view.}
%         \label{fig:Ablation_Curve}
%     \end{figure}

    % First row: LMGS (w/ Both), LMGS (w/o PN), LMGS (w/o RPN)
    % Second row: LMLS, GMLS, GMGS

In this section, we first describe the implementation details of PointNorm and then compare PointNorm with state-of-the-art methods on shape classification and part segmentation benchmarks. Finally, we conduct ablation studies to demonstrate the effectiveness of the proposed modules and the reason for the current hyperparameter selection.

\subsection{Implementation Details}
% Training hyperparams (Scan, Model, Shape)

PointNorm is trained on a single GPU for 200 epochs, using AdamW \cite{loshchilov2017decoupled} as the optimizer and cosine annealing \cite{loshchilov2016sgdr} as the scheduler. The initial learning rate is 0.01; the minimum learning rate is 0.0001; the batch size is 32; the embedding dimension is 64; it takes approximately 9 hours to converge. We utilize both PN and RPN, adopt local mean and global standard deviation, and choose ResBlock with a bottleneck ratio of 1.0. Inspired by previous works \cite{wang2019dynamic, ma2022rethinking}, we use cross-entropy with label smoothing \cite{szegedy2016rethinking} as the loss function. Finally, we adopt random rotation and translation \cite{qi2017pointnet++, wang2019dynamic, thomas2019kpconv, ma2022rethinking} as the data augmentation techniques. 

\begin{table}[t]
\setlength\tabcolsep{4pt}
\caption{Ablation Study Result for PointNorm. FLOPs refer to the floating-point operations for computing a single point cloud. Train/Test Time refers to the seconds per epoch on the training/testing dataset. The best accuracy is in \textbf{Bold}.}
\centering
\begin{tabular}{c|c|cc|cccc}
\hline
                               &                                 & {\begin{tabular}[c]{@{}c@{}}OA \\ (\%) \end{tabular}}       &  {\begin{tabular}[c]{@{}c@{}}mAcc \\ (\%) \end{tabular}}       & FLOPs  & \#Params & \begin{tabular}[c]{@{}c@{}}Train \\ Time \end{tabular} & \begin{tabular}[c]{@{}c@{}}Test \\ Time \end{tabular} \\ \hline
\multirow{3}{*}{\begin{tabular}[c]{@{}c@{}}Layer \\ Number \end{tabular}}  & 24                              & 86.5          & 85.2          & 8.71G  & 7.30M    & 99                                                                  & 9                                                                  \\  
                                        & 40                              & \textbf{86.8} & \textbf{85.6} & 14.59G & 12.63M   & 145                                                                 & 13                                                                 \\ 
                                        & 56                              & 86.7          & 85.0          & 20.48G & 17.95M   & 190                                                                 & 16                                                                 \\ \hline
\multirow{5}{*}{\begin{tabular}[c]{@{}c@{}}Bottleneck \\ Ratio \end{tabular}} & 0.25               & 85.9          & 84.5          & 5.79G  & 4.65M    & 99                                                                  & 9                                                                  \\ 
                                        & 0.50                & 86.6          & 85.4          & 8.72G  & 7.31M    & 111                                                                 & 10                                                                 \\ 
                                        & 1.00                  & \textbf{86.8} & \textbf{85.6} & 14.59G & 12.63M   & 145                                                                 & 13                                                                 \\ 
                                        & 2.00                  & 86.6          & 85.1          & 26.34G & 23.27M   & 205                                                                 & 17                                                                 \\  
                                        % & InvResBlock (2.0)                 & 85.8          & 84.2          & 26.46G & 23.34M   & 252                                                                 & 21                                                                 \\ 
                                        \hline
\multirow{4}{*}{\begin{tabular}[c]{@{}c@{}}Local/ \\ Global \end{tabular}}  & LMGS                            & \textbf{86.8} & \textbf{85.6} & 14.59G & 12.63M   & 145                                                                 & 13                                                                 \\  
                                        & LMLS                            & 81.2          & 78.3          & 14.59G & 12.63M   & 145                                                                 & 13                                                                 \\  
                                        & GMLS                            & 25.1          & 16.8          & 14.59G & 12.63M   & 143                                                                 & 13                                                                 \\ 
                                        & GMGS                            & 78.4          & 75.5          & 14.59G & 12.63M   & 143                                                                 & 13                                                                 \\ \hline
\multirow{3}{*}{Norm.} & w/o PN         & 80.7          & 77.8          & 14.59G & 12.63M   & 136                                                                 & 12                                                                 \\ 
                                        & w/o RPN & 85.6          & 84.1          & 14.59G & 12.63M   & 141                                                                 & 12                                                                 \\  
                                        & w/ both                            & \textbf{86.8} & \textbf{85.6} & 14.59G & 12.63M   & 145                                                                 & 13                                                                 \\ \hline
\end{tabular}
\label{tab:Ablation_ScanObjectNN}
\end{table}

\begin{figure}[t]
    \centering
    
    % First row
    \subfigure[LMGS (w/ Both)]{
     \includegraphics[width=2.6cm]{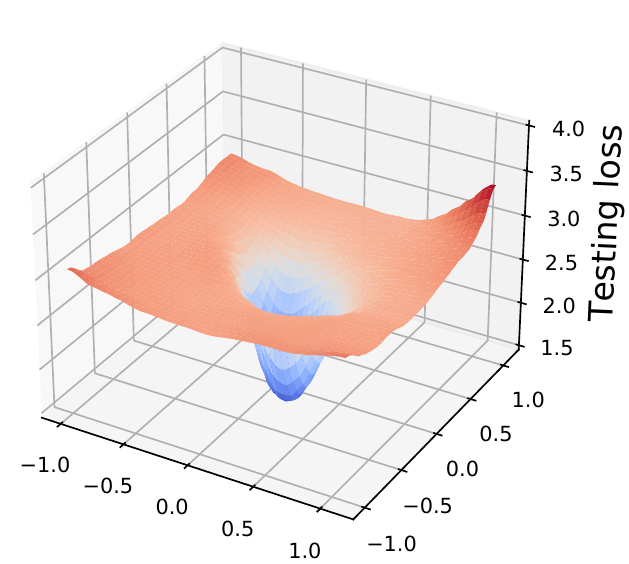}
    }
    \subfigure[LMGS (w/o PN)]{
     \includegraphics[width=2.6cm]{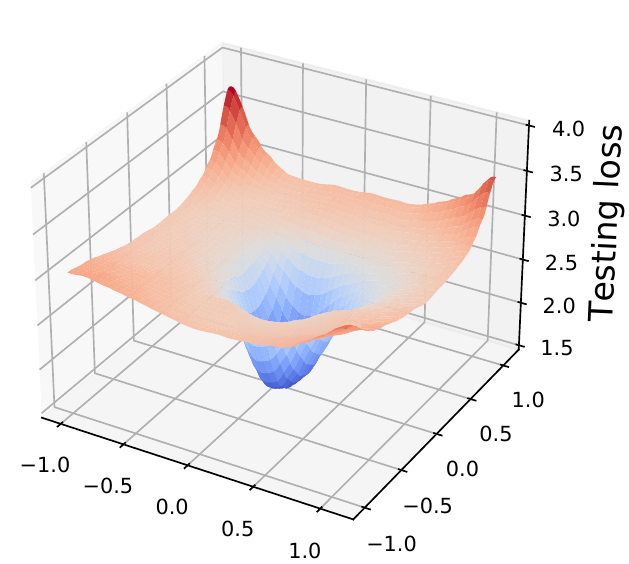}
    }
    \subfigure[LMGS (w/o RPN)]{
     \includegraphics[width=2.6cm]{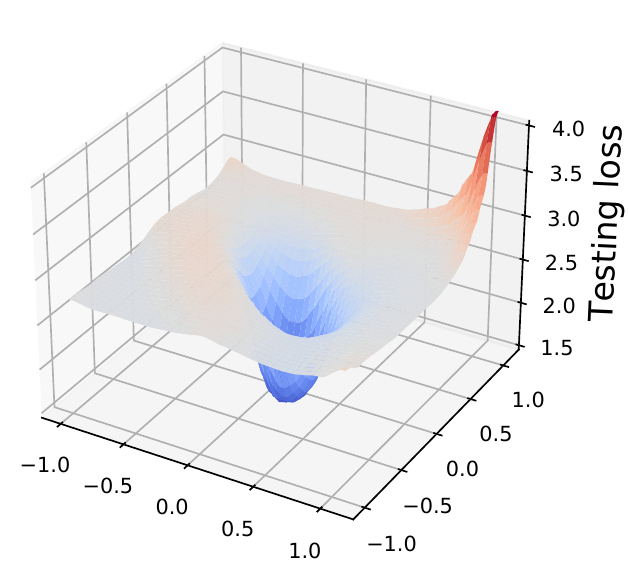}
    } \\

    % Second Row
    \subfigure[LMLS (w/ Both)]{
     \includegraphics[width=2.6cm]{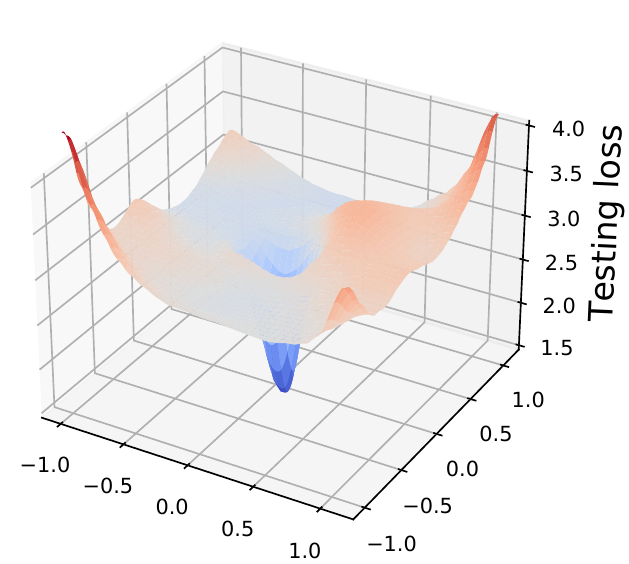}
    }
    \subfigure[GMLS (w/ Both)]{
     \includegraphics[width=2.6cm]{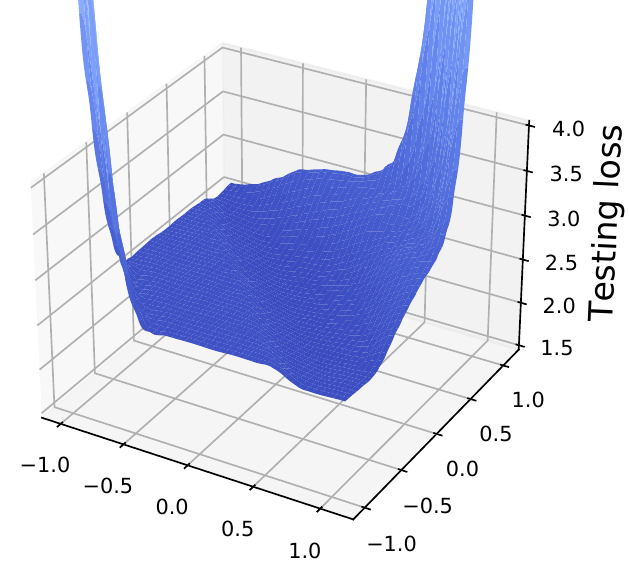}
    }
    \subfigure[GMGS (w/ Both)]{
     \includegraphics[width=2.6cm]{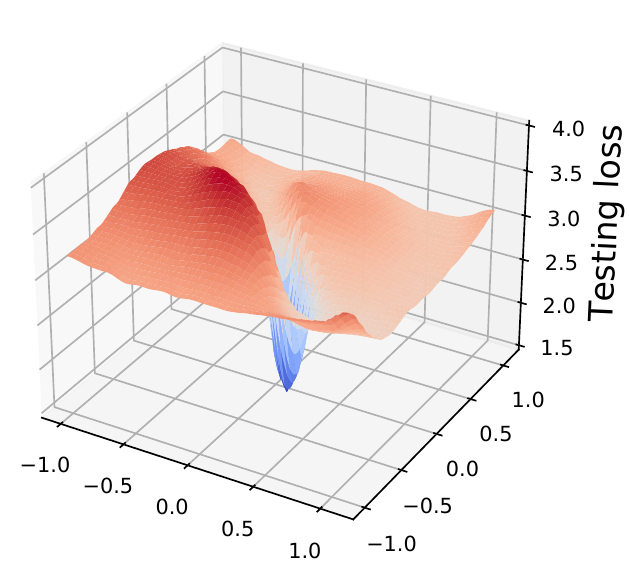}
    }
    \caption{Loss landscape \cite{li2018visualizing} along two random directions for different  PointNorm variants. Zoom in to view the details. }
    \label{fig:Ablation_Land}
\end{figure}

%%% Table for shape classification on ModelNet40 and ScanObjectNN  
\begin{table*}[t]
\small
\setlength\tabcolsep{8pt}
\caption{Shape Classification Result on ModelNet40 and ScanObjectNN. All methods except PointASNL are tested \textbf{without the voting strategy}. We use overall accuracy (OA) and mean accuracy (mAcc), number of parameters (\#Params), train speed and test speed by samples per second computed on a single GPU as the evaluation metrics. We use \textbf{bold} to indicate the best score, and \color{blue}{blue} \color{black}{for the second-best score. `-' indicate a result is unavailable. } }
\centering
\begin{tabular}{c|c|c|cc|cc|ccc}
\hline
                         &                               &                                & \multicolumn{2}{c|}{ModelNet40 \cite{wu20153d}}          & \multicolumn{2}{c|}{ScanObjectNN \cite{uy2019revisiting}}                                              &                              &                               &                              \\ \cline{4-7}
\multirow{-2}{*}{Method} & \multirow{-2}{*}{Publication} & \multirow{-2}{*}{Input} & \multicolumn{1}{c}{OA (\%)} & mAcc (\%) & \multicolumn{1}{c}{OA (\%)}                          & mAcc (\%)                        & \multirow{-2}{*}{\#Params}   & \multirow{-2}{*}{\begin{tabular}[c]{@{}c@{}}Train \\ Speed \end{tabular}} & \multirow{-2}{*}{\begin{tabular}[c]{@{}c@{}}Test \\ Speed \end{tabular}} \\ \hline
PointNet \cite{qi2017pointnet}                 & CVPR 2017                     & 1k                             & \multicolumn{1}{c}{89.2}    & 86.0      & \multicolumn{1}{c}{68.2}                        & 63.4                        & 3.47M                        & -                            & -                           \\ 
PointNet++ \cite{qi2017pointnet++}               & NeurIPS 2017                  & 1k                             & \multicolumn{1}{c}{90.7}    & 88.4      & \multicolumn{1}{c}{77.9}                        & 75.4                        & 1.48M                       & \textbf{223.8}                & {\color{blue}{308.5}} \\ 
PointCNN \cite{li2018pointcnn}                & NeurIPS 2018                  & 1k                             & \multicolumn{1}{c}{92.5}    & 88.1      & \multicolumn{1}{c}{78.5}                        & 75.1                        & -                           & -                            & -                           \\ 
DGCNN \cite{wang2019dynamic}                    & TOG 2019                      & 1k                             & \multicolumn{1}{c}{92.9}    & 90.2      & \multicolumn{1}{c}{78.1}                        & 73.6                        & 1.82M                        & -                            & -                           \\
RS-CNN \cite{liu2019relation}                   & CVPR 2019                     & 1k                             & \multicolumn{1}{c}{92.9}    & -        & \multicolumn{1}{c}{-}                          & -                          & 2.38M                        & -                            & -                           \\ 
PointConv \cite{wu2019pointconv}                & CVPR 2019                     & 1k                             & \multicolumn{1}{c}{92.5}    & -        & \multicolumn{1}{c}{-}                          & -                          & 18.6M                        & 17.9                          & 10.2                         \\ 
KPConv \cite{thomas2019kpconv}                  & ICCV 2019                     & 7k                             & \multicolumn{1}{c}{92.9}    & -        & \multicolumn{1}{c}{-}                          & -                          & 14.3M                        & 31.0                          & 80.0                         \\ 
PointASNL \cite{yan2020pointasnl}           & CVPR 2020                     & 1k                             & \multicolumn{1}{c}{93.2}    & -        & \multicolumn{1}{c}{-}                          & -                          & 10.1M                        & -                            & -                           \\ 
Grid-GCN \cite{xu2020grid}                 & CVPR 2020                     & 1k                             & \multicolumn{1}{c}{93.1}    & \textbf{91.3}      & \multicolumn{1}{c}{-}                          & -                          & -                           & -                            & -                           \\ 
DRNet \cite{qiu2021dense}                    & WACV 2021                     & 1k                             & \multicolumn{1}{c}{93.1}    & -        & \multicolumn{1}{c}{80.3}                        & 78.0                        & -                           & -                            & -                           \\ 
PAConv \cite{xu2021paconv}                   & CVPR 2021                     & 1k                             & \multicolumn{1}{c}{93.6}    & -        & \multicolumn{1}{c}{-}                          & -                          & 2.44M                        & -                            & -                           \\ 
CurveNet \cite{xiang2021walk}                & ICCV 2021                     & 1k                             & \multicolumn{1}{c}{\color{blue}{93.8}}    & -        & \multicolumn{1}{c}{-}                          & -                          & 2.04M                        & 20.8                          & 15.0                         \\ 
GDANet \cite{xu2021learning}                  & AAAI 2021                     & 1k                             & \multicolumn{1}{c}{93.4}    & -        & \multicolumn{1}{c}{-}                          & -                          & {\color{blue}{0.93M}} & 26.3                          & 14.0                         \\
PRANet \cite{cheng2021net}                  & TIP 2021                      & 1k                             & \multicolumn{1}{c}{93.2}    & \color{blue}{90.6}      & \multicolumn{1}{c}{81.0}                          & 77.9                        & -                           & -                            & -                           \\ 
PointMLP \cite{ma2022rethinking}                & ICLR 2022                     & 1k                             & \multicolumn{1}{c}{\textbf{94.1}}    & \textbf{91.3}      & \multicolumn{1}{c}{{\color{blue}{85.4}}} & {\color{blue}{83.9}} & 12.60M                       & 47.1                          & 112.0                        \\
RepSurf-U \cite{ran2022surface}              & CVPR 2022                     & 1k                             & \multicolumn{1}{c}{-}      & -        & \multicolumn{1}{c}{84.6}                        & 81.9                        & 1.48M                        & -                            & -                           \\ \hline
PointNorm                     &                               & 1k                             & \multicolumn{1}{c}{\textbf{94.1}}    & \textbf{91.3}      & \multicolumn{1}{c}{\textbf{86.8}}               & \textbf{85.6}               & 12.63M                       & 58.2                          & 140.0                          \\ 
PointNorm-Tiny              &                               & 1k                             & \multicolumn{1}{c}{93.5}    & \color{blue}{90.6}      & \multicolumn{1}{c}{85.3}                        & 83.6                        & \textbf{0.68M}               & {\color{blue}{196.4}}  & \textbf{420.0}                 \\ \hline
\end{tabular}
\label{tab:Shape_Class}
\end{table*}

% \footnote{PointMLP's result on ModelNet40 dataset are taken from PointMLP's official code repository: https://github.com/ma-xu/pointMLP-pytorch/issues/1. RepSurf report accuracy on ModelNet40 dataset but does not provide code or pretrained model for ModelNet40.}

%% Table for ShapeNetPart Part Segmentation
\begin{table*}[t]
\setlength\tabcolsep{4pt}
\caption{Part Segmentation Result on ShapeNetPart. All scores are reported in \%. The best score for Instance mIoU and Class mIoU are in \textbf{bold}. Zoom in to view better.}
\centering
\scriptsize
\begin{tabular}{c|cc|cccccccccccccccc}
\hline
\textbf{Method} & \textbf{\begin{tabular}[c]{@{}c@{}}Inst. \\ mIoU\end{tabular}} & \textbf{\begin{tabular}[c]{@{}c@{}}Cls. \\ mIoU\end{tabular}} & \textbf{\begin{tabular}[c]{@{}c@{}}air- \\ plane\end{tabular}} & \textbf{bag} & \textbf{cap} & \textbf{car} & \textbf{chair} & \textbf{\begin{tabular}[c]{@{}c@{}}aero- \\ phone\end{tabular}} & \textbf{guitar} & \textbf{knife} & \textbf{lamp} & \textbf{laptop} & \textbf{\begin{tabular}[c]{@{}c@{}}motor- \\ bike\end{tabular}} & \textbf{mug} & \textbf{pistol} & \textbf{rocket} & \textbf{\begin{tabular}[c]{@{}c@{}}skate- \\ board\end{tabular}} & \textbf{table} \\ \hline
PointNet \cite{qi2017pointnet}        & 83.7                                                           & 80.4                                                          & 83.4              & 78.7         & 82.5         & 74.9         & 89.6           & 73.0              & 91.5            & 85.9           & 80.8          & 95.3            & 65.2               & 93.0         & 81.2            & 57.9            & 72.8                & 80.6           \\ 
PointNet++ \cite{qi2017pointnet++}      & 85.1                                                           & 81.9                                                          & 82.4              & 79.0         & 87.7         & 77.3         & 90.8           & 71.8              & 91.0            & 85.9           & 83.7          & 95.3            & 71.6               & 94.1         & 81.3            & 58.7            & 76.4                & 82.6           \\ 
PCNN \cite{atzmon2018point}           & 85.1                                                           & 81.8                                                          & 82.4              & 80.1         & 85.5         & 79.5         & 90.8           & 73.2              & 91.3            & 86.0           & 85.0          & 95.7            & 73.2               & 94.8         & 83.3            & 51.0            & 75.0                & 81.8           \\ 
DGCNN \cite{wang2019dynamic}           & 85.2                                                           & 82.3                                                          & 84.0              & 83.4         & 86.7         & 77.8         & 90.6           & 74.7              & 91.2            & 87.5           & 82.8          & 95.7            & 66.3               & 94.9         & 81.1            & 63.5            & 74.5                & 82.6           \\
PointCNN \cite{li2018pointcnn}        & 86.1                                                           & 84.6                                                          & 84.1              & 86.5         & 86.0         & 80.8         & 90.6           & 79.7              & 92.3            & 88.4           & 85.3          & 96.1            & 77.2               & 95.2         & 84.2            & 64.2            & 80.0                & 83.0           \\
RS-CNN \cite{liu2019relation}          & \textbf{86.2}                                                           & 84.0                                                          & 83.5              & 84.8         & 88.8         & 79.6         & 91.2           & 81.1              & 91.6            & 88.4           & 86.0          & 96.0            & 73.7               & 94.1         & 83.4            & 60.5            & 77.7                & 83.6           \\
SyncSpecCNN \cite{yi2017syncspeccnn}        & 84.7                                                           & 82.0                                                          & 81.6              & 81.7         & 81.9         & 75.2         & 90.2           & 74.9              & 93.0            & 86.1           & 84.7          & 95.6            & 66.7               & 92.7         & 81.6            & 60.6            & 82.9                & 82.1           \\ 
SPLATNet \cite{su2018splatnet}       & 85.4                                                           & 83.7                                                          & 83.2              & 84.3         & 89.1         & 80.3         & 90.7           & 75.5              & 92.1            & 87.1           & 83.9          & 96.3            & 75.6               & 95.8         & 83.8            & 64.0            & 75.5                & 81.8           \\ 
SpiderCNN \cite{xu2018spidercnn}       & 85.3                                                           & 82.4                                                          & 83.5              & 81.0         & 87.2         & 77.5         & 90.7           & 76.8              & 91.1            & 87.3           & 83.3          & 95.8            & 70.2               & 93.5         & 82.7            & 59.7            & 75.8                & 82.8           \\
PAConv \cite{xu2021paconv}          & 86.1                                                           & 84.6                                                          & 84.3              & 85.0         & 90.4         & 79.7         & 90.6           & 80.8              & 92.0            & 88.7           & 82.2          & 95.9            & 73.9               & 94.7         & 84.7            & 65.9            & 81.4                & 84.0           \\ 
PointMLP \cite{ma2022rethinking}        & 86.1                                                           & 84.6                                                          & 83.5              & 83.4         & 87.5         & 80.5         & 90.3           & 78.2              & 92.2            & 88.1           & 82.6          & 96.2            & 77.5               & 95.8         & 85.4            & 64.6            & 83.3                & 84.3           \\ \hline
PointNorm            & \textbf{86.2}                                                           & \textbf{84.7}                                                          & 82.7              & 84.9         & 88.9         & 79.8         & 90.2           & 81.9             & 91.6            & 87.4           & 82.9          & 95.8            & 78.4               & 95.5         & 84.5            & 65.6            & 81.4                & 83.8           \\
PointNorm-Tiny      & 85.6                                                           & 84.5                                                          & 82.9              & 88.0         & 89.7         & 79.3         & 90.1           & 79.9              & 91.6            & 87.7           & 82.4          & 95.8            & 76.3               & 95.0         & 83.5            & 64.6            & 81.9                & 83.5           \\ \hline
\end{tabular}
\label{tab:Shape_Part}
\end{table*}

\subsection{Ablation Studies} \label{Ablation Studies}
We conduct extensive ablation studies on the ScanObjectNN dataset to demonstrate the effectiveness of the proposed components in PointNorm, and the reason for specific module designs and hyperparameters selections. Specifically, we employ a score table (Table \ref{tab:Ablation_ScanObjectNN}) and loss landscape visualizations (Fig. \ref{fig:Ablation_Land}) to investigate four aspects, including Layer Number, Bottleneck Ratio, Local/Global, and Normalization.

\subsubsection{Layer Number}
We define the Layer Number as the number of learnable layers in a network except for batch normalization and activation functions. Table \ref{tab:Ablation_ScanObjectNN} shows that the 40-layer variant gives the best OA and mAcc. That being said, all Layer Number variants of PointNorm (24-layer, 40-layer, 56-layer) have a close OA and mAcc. That demonstrates the robustness of PointNorm to Layer Number in the network. 

% Indeed, we can easily adjust the Layer Number by changing the numbers of Resblock in PointNorm.

% Besides, the first column of Fig. \ref{fig:Ablation_Curve} displays that the 40-layer variant has the most consistent accuracy growth and the most stable model convergence. 

% All Layer Number variants of PointNorm (24-layer, 40-layer, 56-layer) have a close OA and mAcc, much better than the previous state-of-the-art. That demonstrates the robustness of PointNorm to Layer Number in the network. 

\subsubsection{Bottleneck Ratio}
The Bottleneck Ratio of Residual Block \cite{he2016deep} offers a convenient trade-off between model size and performance. We consider four bottleneck ratios (0.25, 0.50, 1.00, 2.00) to analyze PointNorm's performance at different sizes. Table \ref{tab:Ablation_ScanObjectNN} shows that a Bottleneck ratio of 1.00 is the best choice because it gives the best OA and mAcc. 

% sacrifice OA and mAcc for boost in XXX and XXX. 

% We additionally add InvResBlock with a bottleneck ratio of 2.0 to aid comparison.

% Besides, we find that ResBlock performs better than InvResBlock, under the same or even with smaller bottleneck ratios. For example, ResBlock (2.0) surpasses InvResBlock (2.0) by 0.8\% in OA and 0.9\% in mAcc, while still having better FLOPS, Params, Train Time, and Test Time. ResBlock's superiority is more evident in the second column of Fig. \ref{fig:Ablation_Curve}. With approximately 22\% FLOPs and 20\% Params, ResBlock (0.25) outperforms InvResBlock (2.0) for both OA and mAcc towards the end of training. 

% Maybe: explore other variants (e.g., 0.75, 1.25, 1.5)

% We consider two choices for calculating our the mean and the standard deviation in DualNorm. The first choice is \textit{Local}, where the computation is done within each group (i.e., the KNN-grouped neighbor points) in a training batch. The second choice is \textit{Global}, where the computation is done for all points in a training batch. 

\subsubsection{Local/Global}
% result table and loss landscape
As mentioned in Subsection \ref{DualNorm}, we can choose either \textit{Local} or \textit{Global} for calculating the mean and the standard deviation. Therefore, there exist four combinations, including Local Mean Global Standard Deviation (LMGS), Local Mean Local Standard Deviation (LMLS), Global Mean Local Standard Deviation (GMLS), and Global Mean Global Standard Deviation (GMGS). Table \ref{tab:Ablation_ScanObjectNN} gives us the quick answer: LMGS is the optimal choice.

We visualize the loss landscape in Fig. \ref{fig:Ablation_Land} to better understand the underlying mechanism of DualNorm. We notice that both LMLS and GMGS have sharp minimas and rough surfaces with many hills, which is hard to train. GMLS can hardly be optimized because it has no apparent minima. In comparison, LMGS has a flat minima and a smooth landscape, which remarkably facilities the optimization process \cite{keskar2016large, li2018visualizing}. Based on the analysis above, we confirm that LMGS is the best. 

% % result table
% Table \ref{tab:Ablation_ScanObjectNN} show that GMLS's OA and mAcc stuck at a low score and that GMGS's OA and mAcc have significant fluctuations. In comparison, LMLS and LMGS have more consistent growth patterns for OA and mAcc. However, LMGS has remarkably better scores for both OA and mAcc. 

\subsubsection{Normalization}
% brief explanation 
We investigate the building blocks of DualNorm (PN, RPN) by considering three variants: PointNorm (w/o PN, w/o RPN, w/ both PN and RPN). 

% result from table and loss landscape
Table \ref{tab:Ablation_ScanObjectNN} shows that PointNorm w/ Both PN and RPN has the best OA and mAcc. Fig. \ref{fig:Ablation_Land} shows that PointNorm w/ both PN and RPN has a flat minima and a smooth surface. While PointNorm w/o PN and PointNorm w/o RPN also have flat minima, they have significant uphills at the border, which may lead to loss explorations. Hence, we conclude that both PN and RPN are essential building blocks of DualNorm.

\subsection{Shape Classification}
    % dataset: ModelNet40 (Synthetic), ScanObjectNN (Real-world)
    % metrics: mAcc, OA

We evaluate PointNorm on shape classification tasks using two benchmark datasets. The first benchmark is ModelNet40 \cite{wu20153d}, a synthetic dataset with 40 classes, 9,843 training samples, and 581 testing samples. The second benchmark is ScanobjectNN \cite{uy2019revisiting}, a real-world dataset with 15 classes, 2,321 training samples, and 581 testing samples.  

We show the experiment results for shape classification in Table \ref{tab:Shape_Class} \footnote{RepSurf reports accuracy on both ModelNet40 and ScanObjectNN. However, its code and pre-trained model for ModelNet40 are missing. Therefore, we report RepSurf's result on ModelNet40 as unavailable.}. For both ModelNet40 and ScanObjectNN, PointNorm has the best OA and mAcc. Notably, on ScanObjectNN, PointNorm surpasses the previous state-of-the-art PointMLP by 1.4\% in OA and 1.7\% in mAcc. The lightweight version PointNorm-Tiny has the smallest \#Params and the best testing speed yet still delivers strong results, surpassing recent state-of-the-arts like PRANet \cite{cheng2021net} and DRNet \cite{qiu2021dense}.

% how about PointNorm-Tiny

% On ScanObjectNN dataset, PointNorm has the best OA of 86.8\% and the best mAcc of 85.6\%. Notably, PointNorm surpasses the recent state-of-the-art PointMLP by 1.4\% for OA and 1.7\% for mAcc. On ModelNet40 dataset, PointNorm has the best mAcc of 91.3\% and the second-best OA of 93.7\%, only 0.1\% less than the recent state-of-the-art CurveNet (93.8\%). However, we note that the train speed of PointNorm (58.2 samples/second) is approximately 2 times faster than CurveNet (20.8 samples/second), and the test speed of PointNorm (140.0 samples/second) is more than 8 times faster than CurveNet (15.0 samples/second). In point cloud analysis, where computational efficiency is of significant priority \cite{guo2020deep}, The clear advantage in train and test speed should be more critical than the 0.1\% gap for OA.

% As mentioned in Subsection \ref{PointNorm-Tiny}, the introduction of PointNorm-Tiny aims to enhance the training and testing speed and, more importantly, reduce the number of parameters. It is worth mentioning that PointNorm-Tiny has the smallest number of parameters (0.68M), the best test speed (420.0 samples/second), and the second-best train speed (196.4 samples/second). Although it is lightweight and computationally efficient, PointNorm-Tiny delivers promising results on ModelNet40 and ScanObjectNN, surpassing recent state-of-the-art methods like PRANet \cite{cheng2021net} and DRNet \cite{qiu2021dense}.

% PointASNL is the only method in the table that use the voting strategy (because its result without voting are unavailable).

\begin{table}[t]
\setlength\tabcolsep{2.5pt}
\caption{Semantic Segmentation Result on S3DIS Dataset. mIOU, mAcc and OA are reported in \%. The best score is in \textbf{Bold}. `-' indicate a result is unavailable.}
\centering
\begin{tabular}{c|ccc|ccc|cc}
\hline
\multirow{2}{*}{Method}                                              & \multicolumn{3}{c|}{S3DIS 6-Fold}                            & \multicolumn{3}{c|}{S3DIS Area-5}           &   \multirow{2}{*}{\#Params}   & \multirow{2}{*}{FLOPs}            \\ \cline{2-7} 
                                                                     & \multicolumn{1}{c}{mIoU} & \multicolumn{1}{c}{mAcc} & OA   & \multicolumn{1}{c}{mIoU} & \multicolumn{1}{c}{mAcc} & OA     \\ \hline
PointNet \cite{qi2017pointnet}                                                             & \multicolumn{1}{c}{47.6} & \multicolumn{1}{c}{66.2} & 78.5 & \multicolumn{1}{c}{43.2} & \multicolumn{1}{c}{52.6} & 77.8 & 1.7M  & 4.1G    \\ 
PointWeb \cite{zhao2019pointweb}                                                             & \multicolumn{1}{c}{66.7} & \multicolumn{1}{c}{76.2} & 87.3 & \multicolumn{1}{c}{60.2} & \multicolumn{1}{c}{66.6} & 87.0 & - & - \\ 
KPConv \cite{thomas2019kpconv}                                                               & \multicolumn{1}{c}{70.6} & \multicolumn{1}{c}{\textbf{79.1}} & -     & \multicolumn{1}{c}{\textbf{67.1}} & \multicolumn{1}{c}{\textbf{72.8}} & -  & 14.9M & -  \\ 
PointASNL \cite{yan2020pointasnl}                                                            & \multicolumn{1}{c}{68.7} & \multicolumn{1}{c}{79.0} & \textbf{88.8} & \multicolumn{1}{c}{62.6} & \multicolumn{1}{c}{68.5} & 87.7 &  22.4M & 19.1G \\ 

% PAConv \cite{xu2021paconv}                                                               & \multicolumn{1}{c}{69.3} & \multicolumn{1}{c}{78.6}     & -
% & \multicolumn{1}{c}{66.5}     & \multicolumn{1}{c}{73.0}     &  - & - &  1.3G \\ 

RPNet \cite{ran2021learning}                                                               & \multicolumn{1}{c}{\textbf{70.8}} & \multicolumn{1}{c}{-}     & -
& \multicolumn{1}{c}{-}     & \multicolumn{1}{c}{-}     &  - & 2.4M &  5.1G \\

DSPoint \cite{zhang2021dspoint}                                                             & \multicolumn{1}{c}{63.3} & \multicolumn{1}{c}{70.9} & -     & \multicolumn{1}{c}{-}     & \multicolumn{1}{c}{-}     &   -  & -  & - \\ \hline

PointNet++ \cite{qi2017pointnet++}                                                           & \multicolumn{1}{c}{54.5} & \multicolumn{1}{c}{67.1} & 81.0 & \multicolumn{1}{c}{52.6} & \multicolumn{1}{c}{63.1} & 82.3  & 0.969M  & 1.00G \\ \hline

\begin{tabular}[c]{@{}c@{}}PointNet++ \\ (w/ DualNorm)\end{tabular} & \multicolumn{1}{c}{62.7}     & \multicolumn{1}{c}{73.8}     &  85.7    & \multicolumn{1}{c}{57.6}     & \multicolumn{1}{c}{68.2}     &  \textbf{88.4}   & 1.006M & 1.05G  \\ 
                                                                      \hline
 (w/ DualNorm)                                                               & \multicolumn{1}{c}{$\uparrow$\color{red}{8.2}} & \multicolumn{1}{c}{$\uparrow$\color{red}{6.7}}     & $\uparrow$\color{red}{4.7}
& \multicolumn{1}{c}{$\uparrow$\color{red}{5.0}}     & \multicolumn{1}{c}{$\uparrow$\color{red}{5.1}}     &  $\uparrow$\color{red}{6.1}  & $\uparrow$\color{red}{0.037M} & $\uparrow$\color{red}{0.05G} \\ \hline                                                                     
\end{tabular}
\label{tab:sem_seg}
\end{table}

\subsection{Part Segmentation}
\begin{figure}[t]
        \centering
        
        % Airplane
            \includegraphics[width=2.8cm]{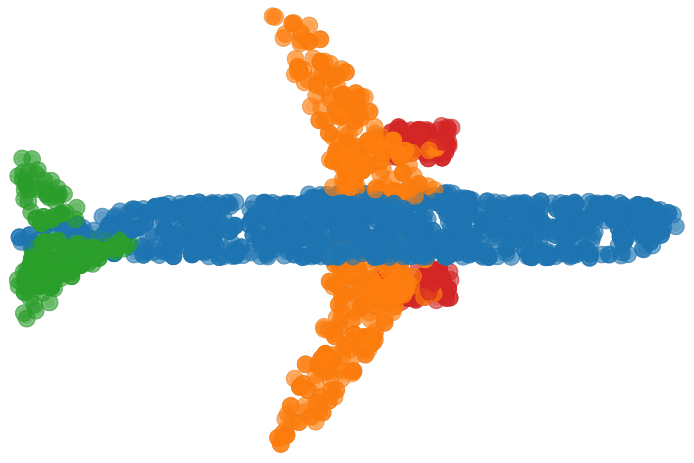}
            \includegraphics[width=2.8cm]{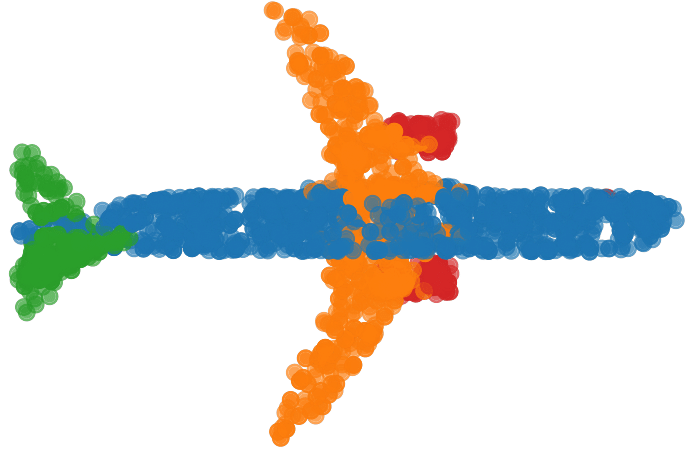}
            \includegraphics[width=2.8cm]{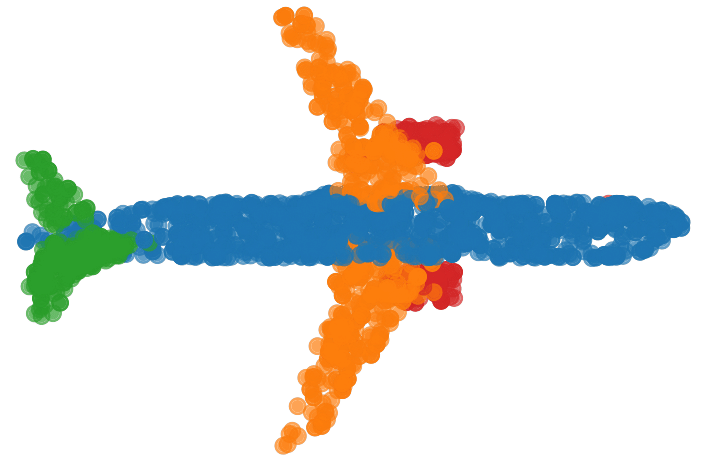} \\

        % Car
            \includegraphics[width=2.8cm]{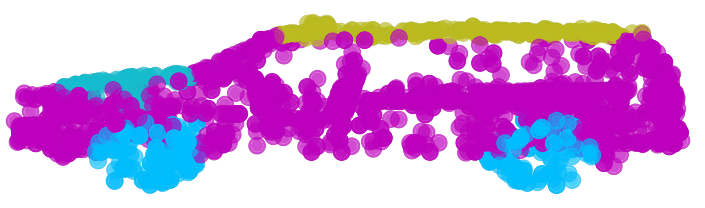}
            \includegraphics[width=2.8cm]{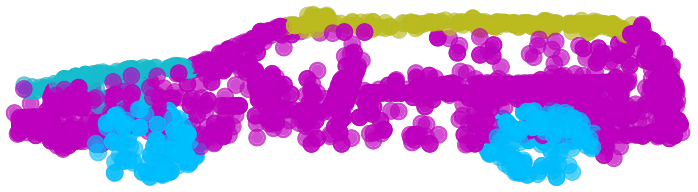}
            \includegraphics[width=2.8cm]{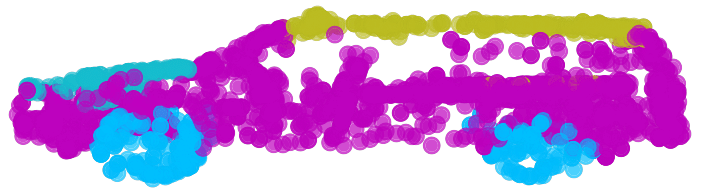} \\
            
        % Chair
            \includegraphics[width=2.8cm]{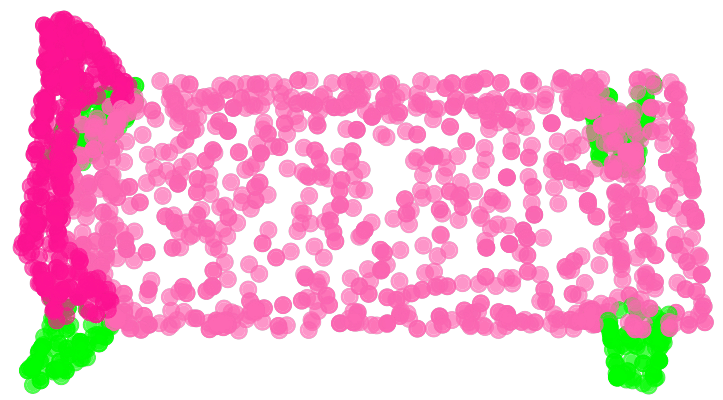}
            \includegraphics[width=2.8cm]{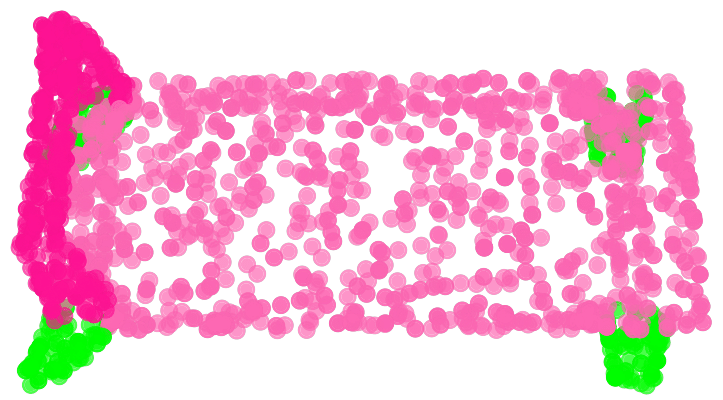}
            \includegraphics[width=2.8cm]{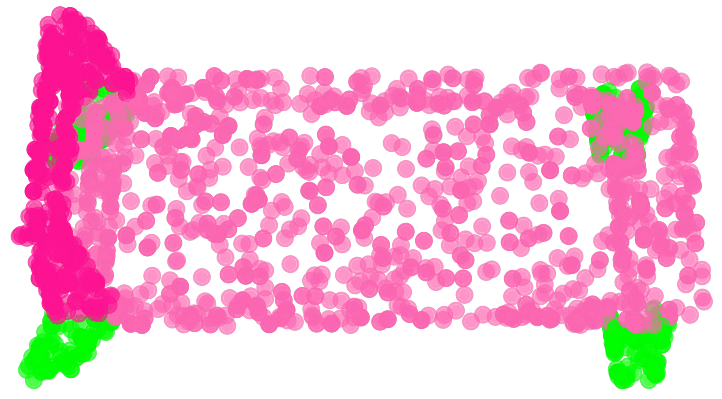} \\
        
        % Lamp
            \includegraphics[width=2.8cm]{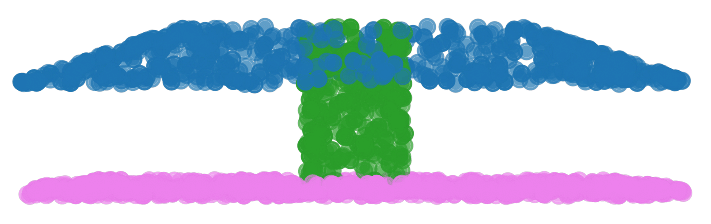}
            \includegraphics[width=2.8cm]{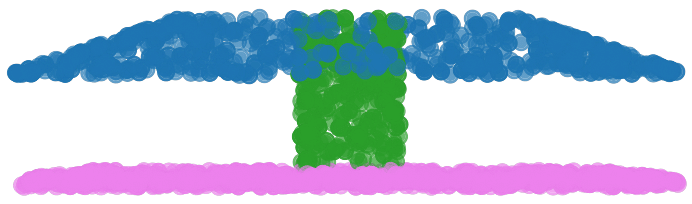}
            \includegraphics[width=2.8cm]{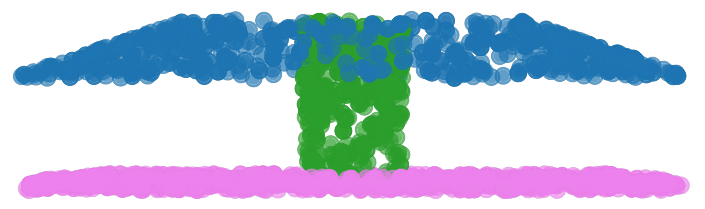} \\
        
        % Skateboard
            \includegraphics[width=2.8cm]{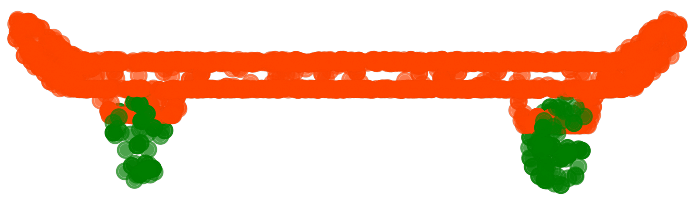}
            \includegraphics[width=2.8cm]{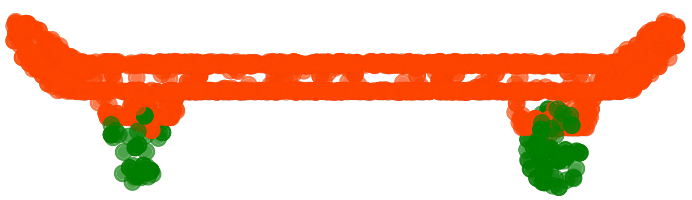}
            \includegraphics[width=2.8cm]{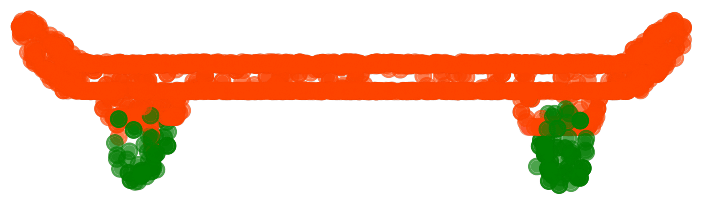} \\
        
        \caption{Part Segmentation Result Visualization. From left to right column: Ground Truth, PointNorm, PointNorm-Tiny. From top to bottom row: airplane, car, chair, lamp, skateboard.}
        \label{fig:Part_Seg}
    \end{figure}

We evaluate our PointNorm on the part segmentation task, using ShapeNetPart \cite{yi2016scalable} as the benchmark dataset. The ShapeNetPart dataset contains 16,881 shapes, 16 classes, and 50 parts labels. The experiment results for part segmentation in Table \ref{tab:Shape_Part} show that PointNorm achieves the best score for both instance mIoU (86.2\%) and class mIoU (84.7\%). PointNorm-Tiny, albeit being lightweight, delivers compelling instance mIoU (85.6\%) and class mIoU (84.5\%), close to the recent state-of-the-arts PAConv \cite{xu2021paconv} and PointMLP \cite{ma2022rethinking}.  We also visualize the part segmentation results in Fig. \ref{fig:Part_Seg}. It can be seen that both PointNorm and PointNorm-Tiny's predictions are incredibly close to the ground truth. 

% Notably, PointNorm-Tiny's class IoU for the bag is 88.0\%, which is the best among all methods.

% More part segmentation results will be in the supplementary material.
        
  % The loss landscape visualization for other PointNorm variants will be in the supplementary material.
    
\subsection{Semantic Segmentation}
\begin{figure}[t]
        \centering
        
        % % Area1
        %     \includegraphics[width=2.5cm]{Sem_Seg/pointnet2/snapshot00.png}
        %     \includegraphics[width=2.5cm]{Sem_Seg/pointnet2_w_D/snapshot00.png}
        %     \includegraphics[width=2.5cm]{Sem_Seg/gt/snapshot00.png}

        % Area2
             \includegraphics[width=2.8cm]{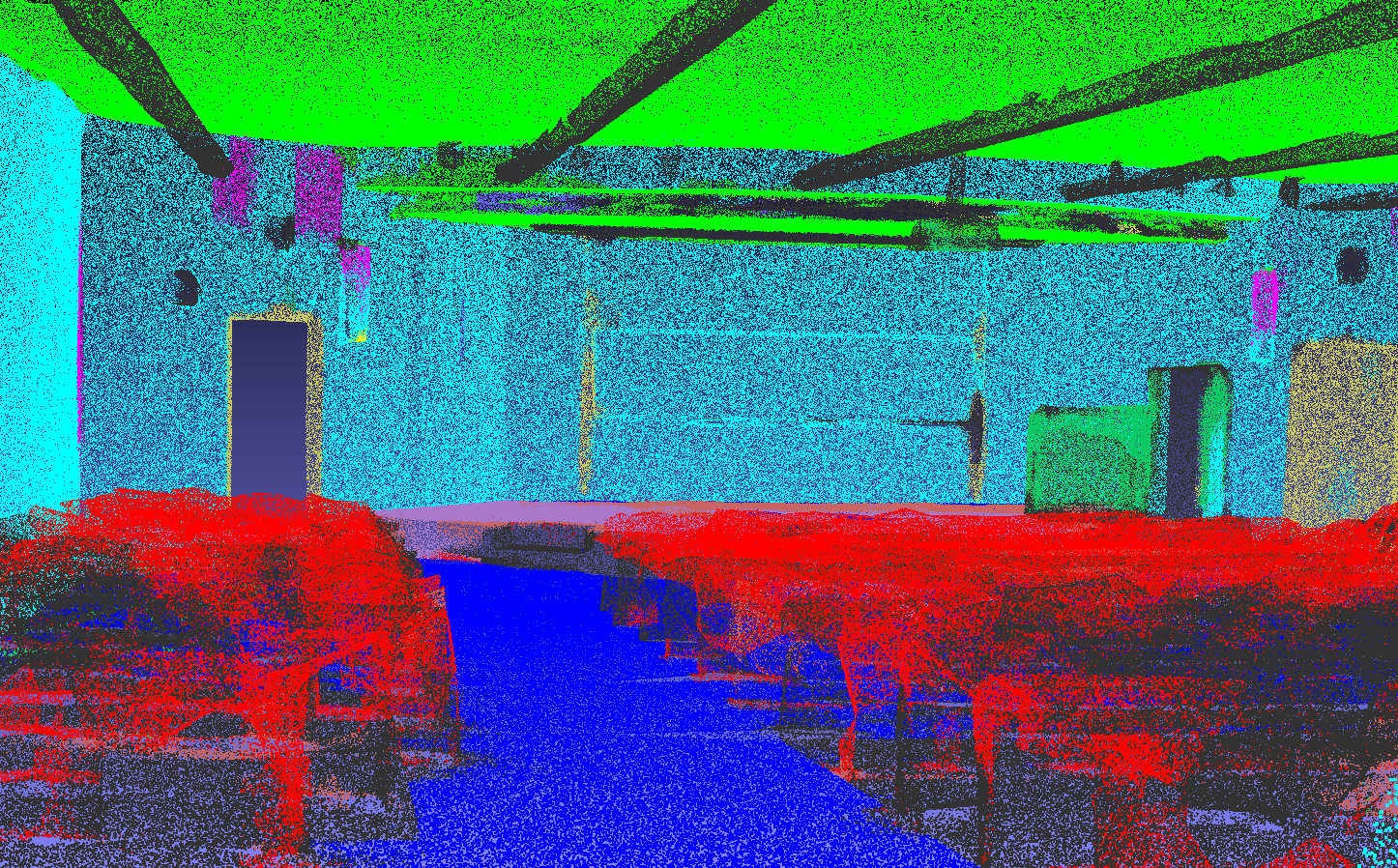}
            \includegraphics[width=2.8cm]{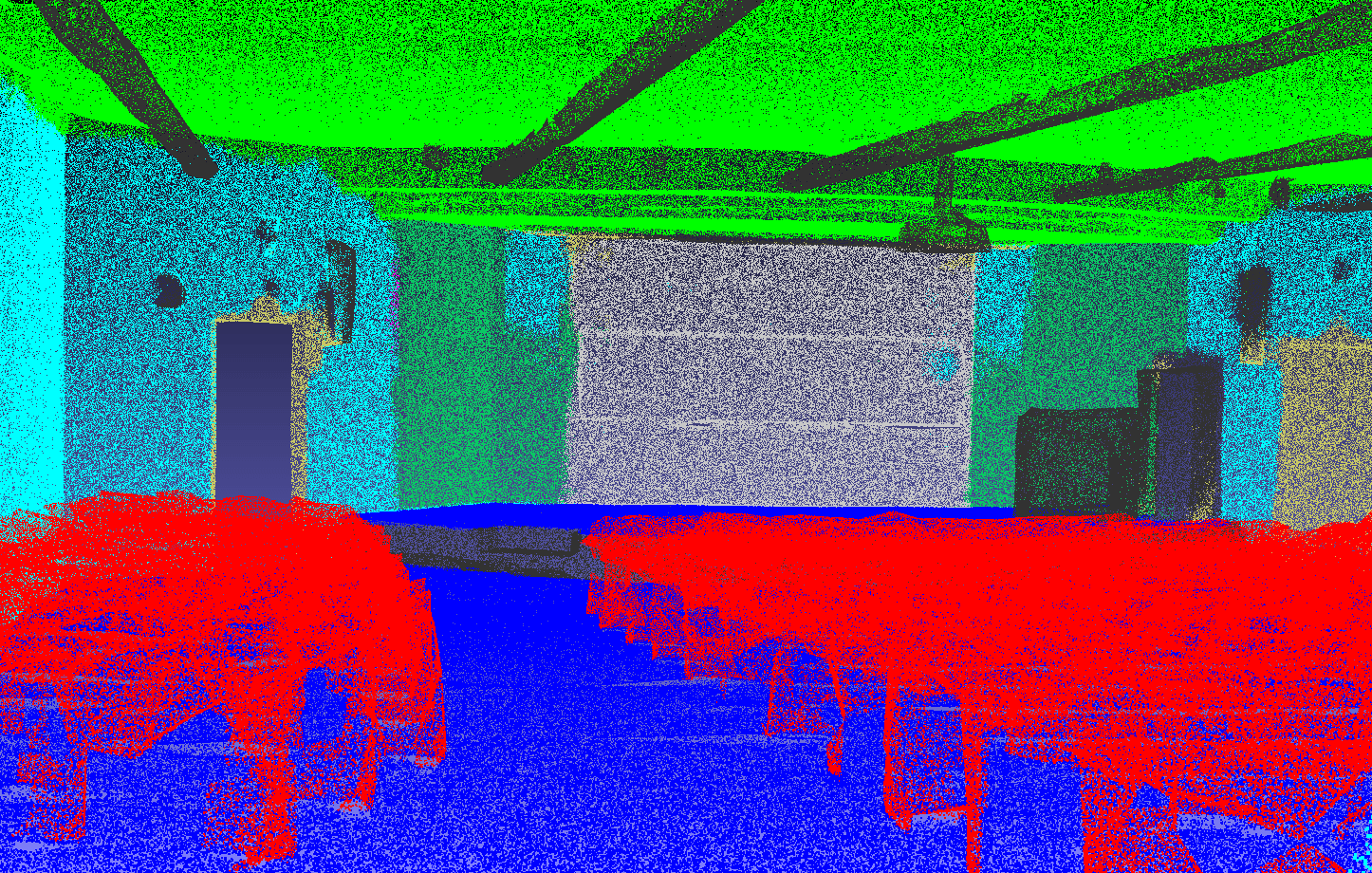}
            \includegraphics[width=2.8cm]{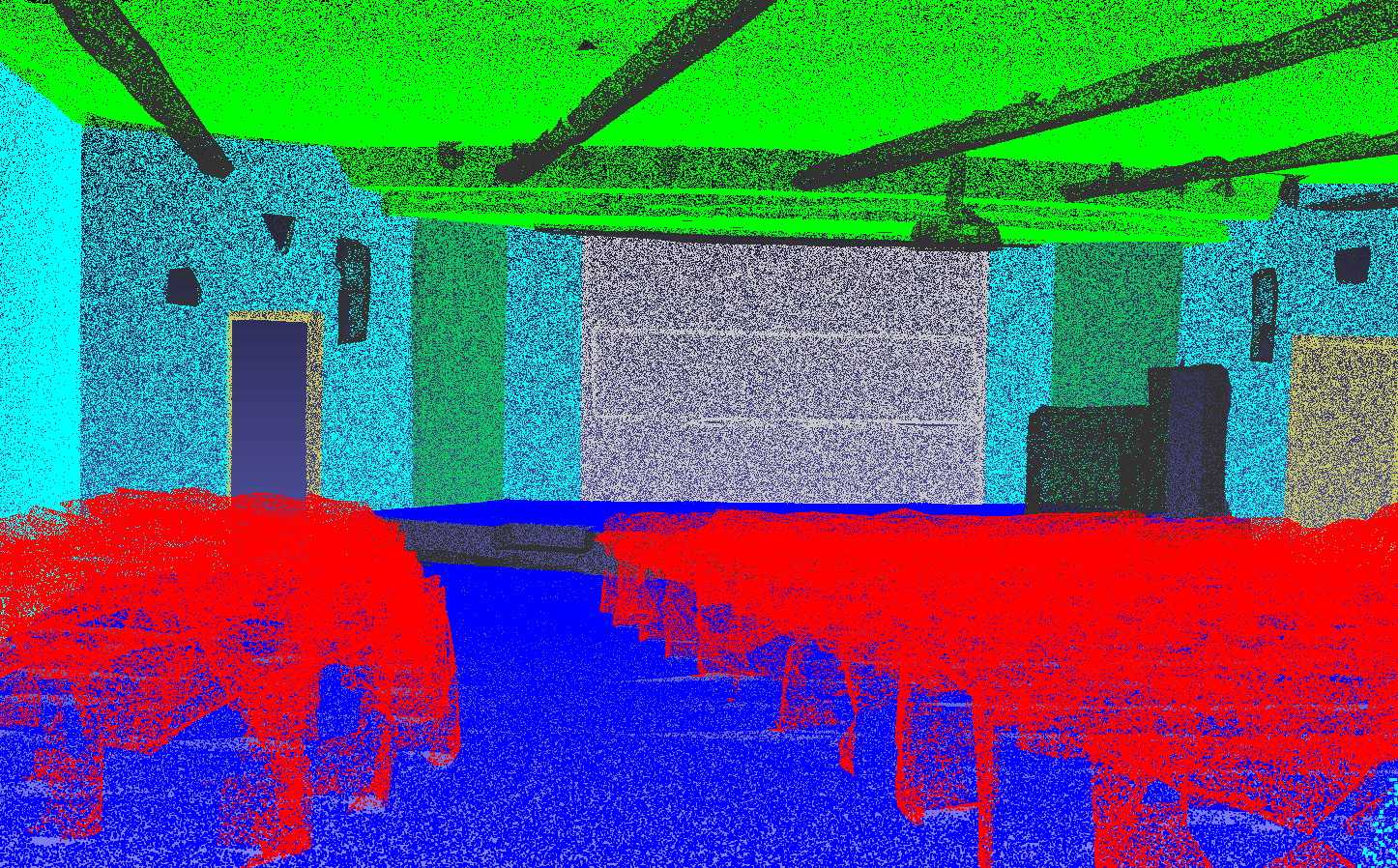}
            
        % Area3
            \includegraphics[width=2.8cm]{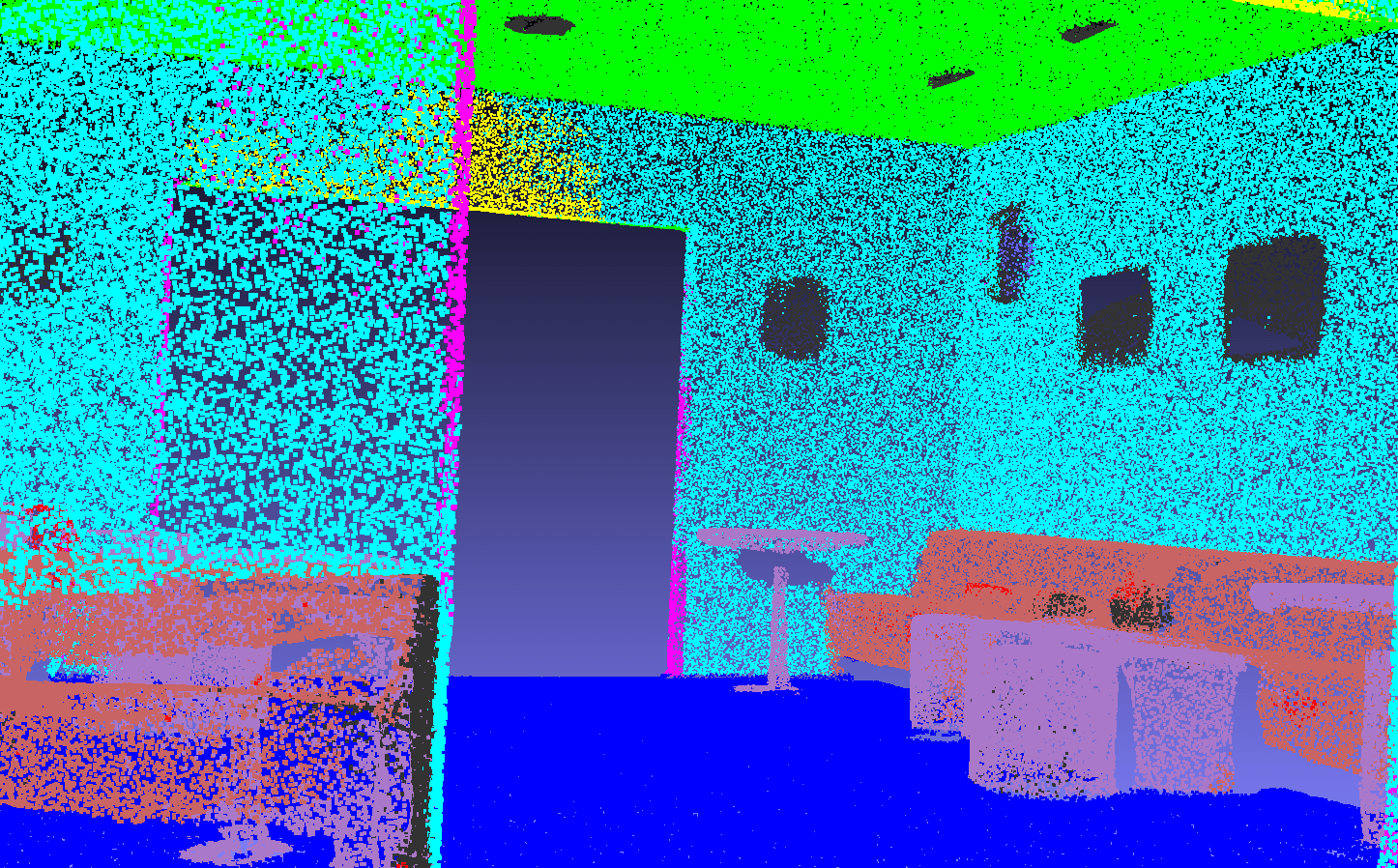}
            \includegraphics[width=2.8cm]{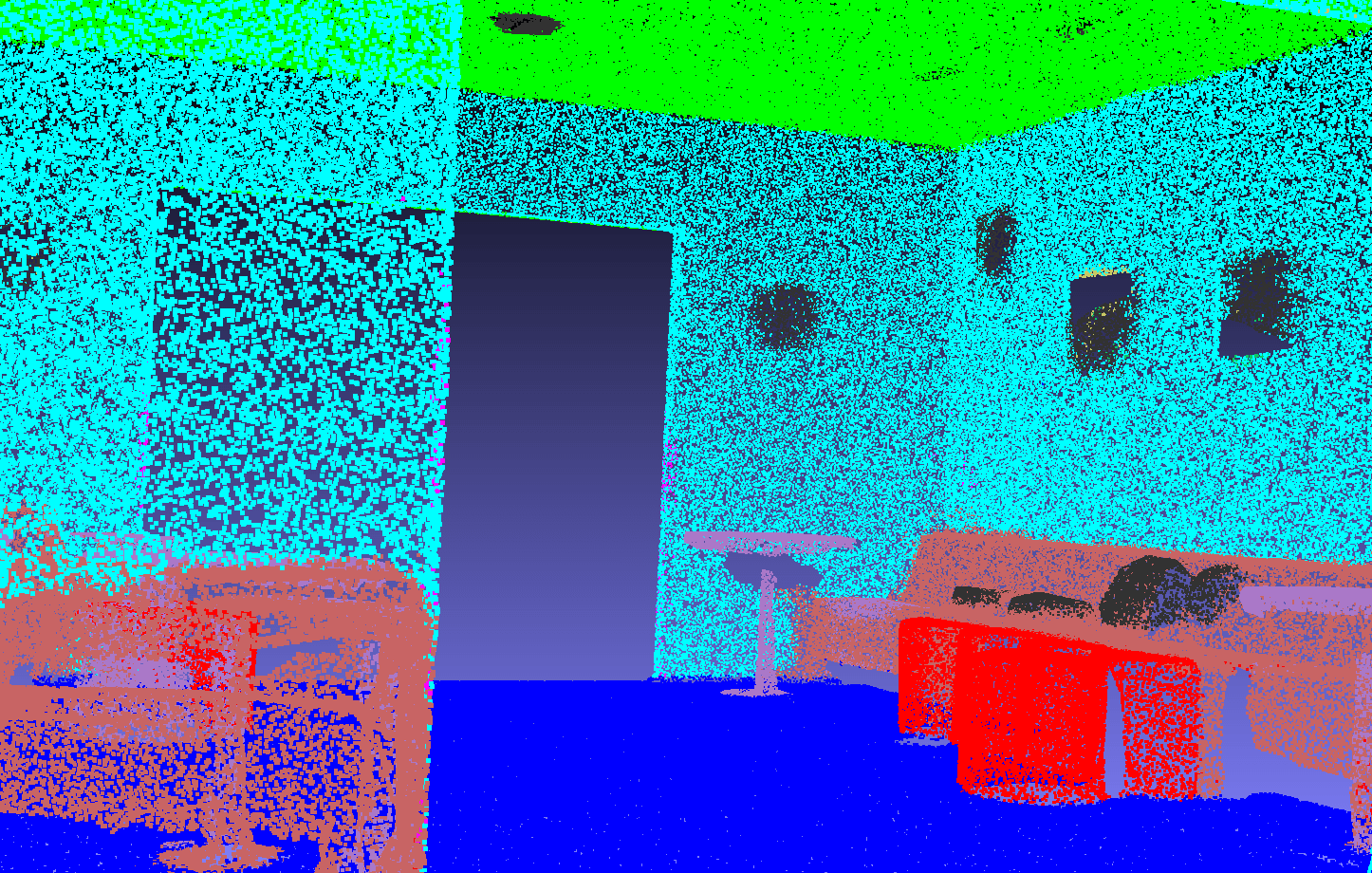}
            \includegraphics[width=2.8cm]{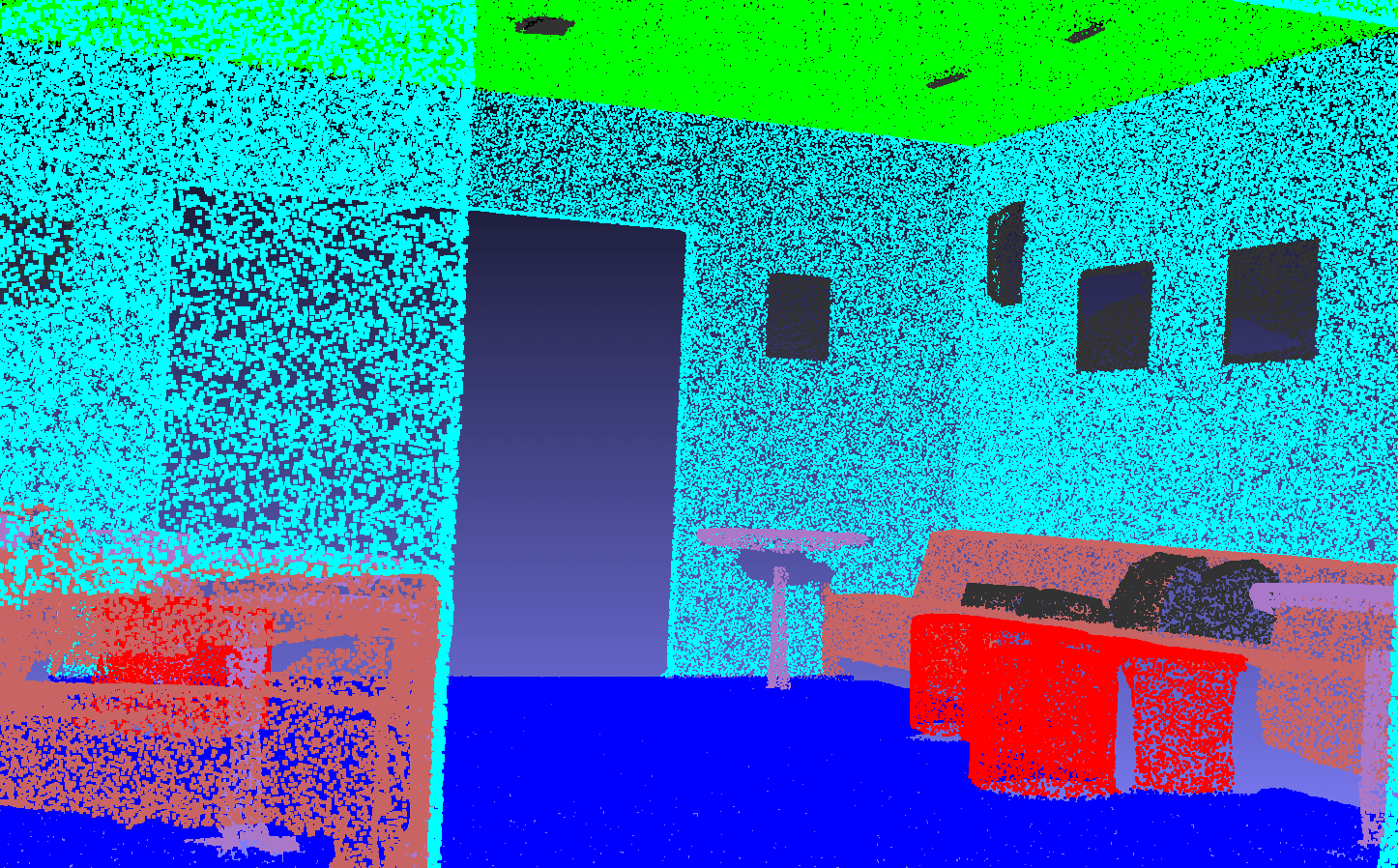}
        
        % Area4
            \includegraphics[width=2.8cm]{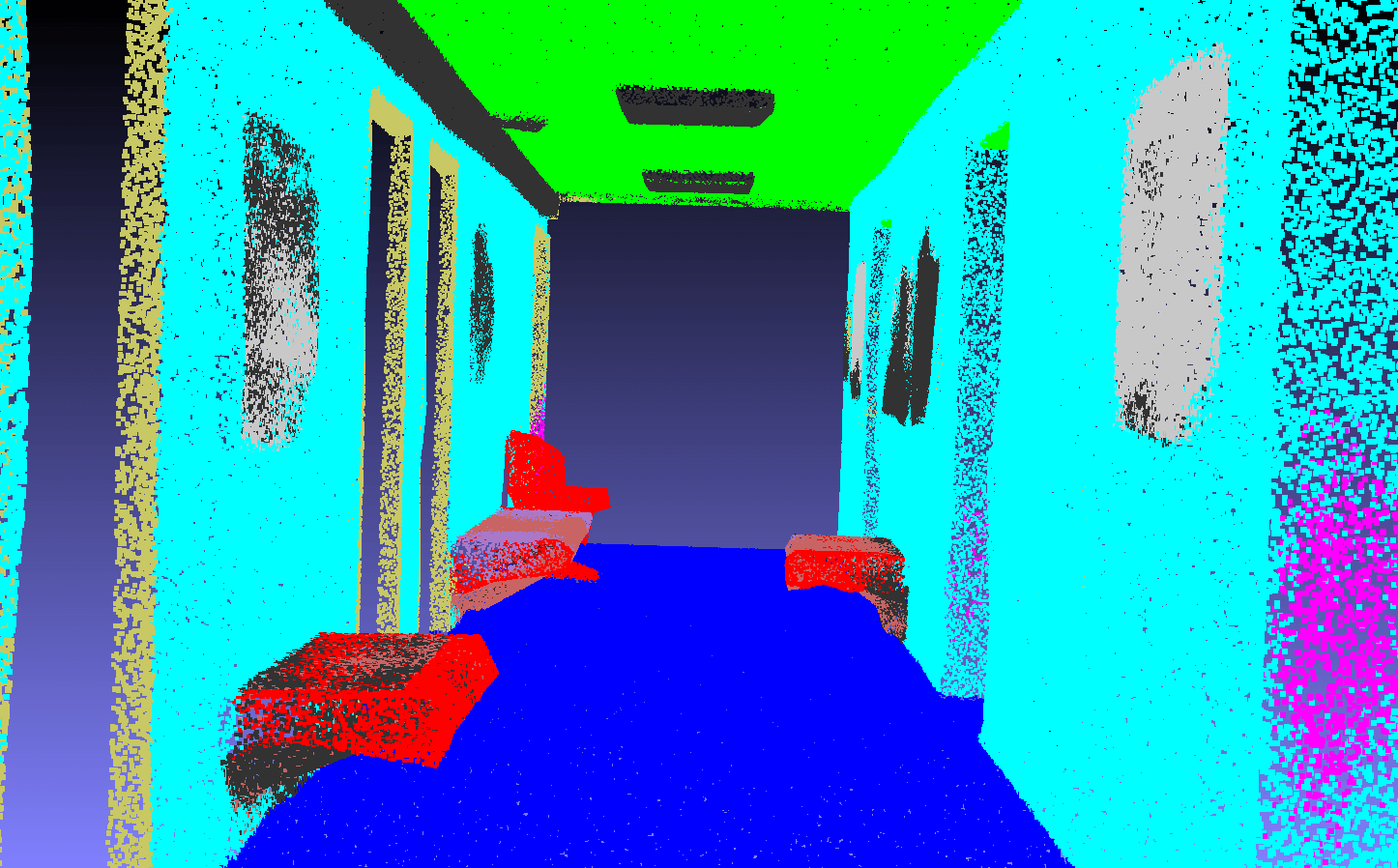}
            \includegraphics[width=2.8cm]{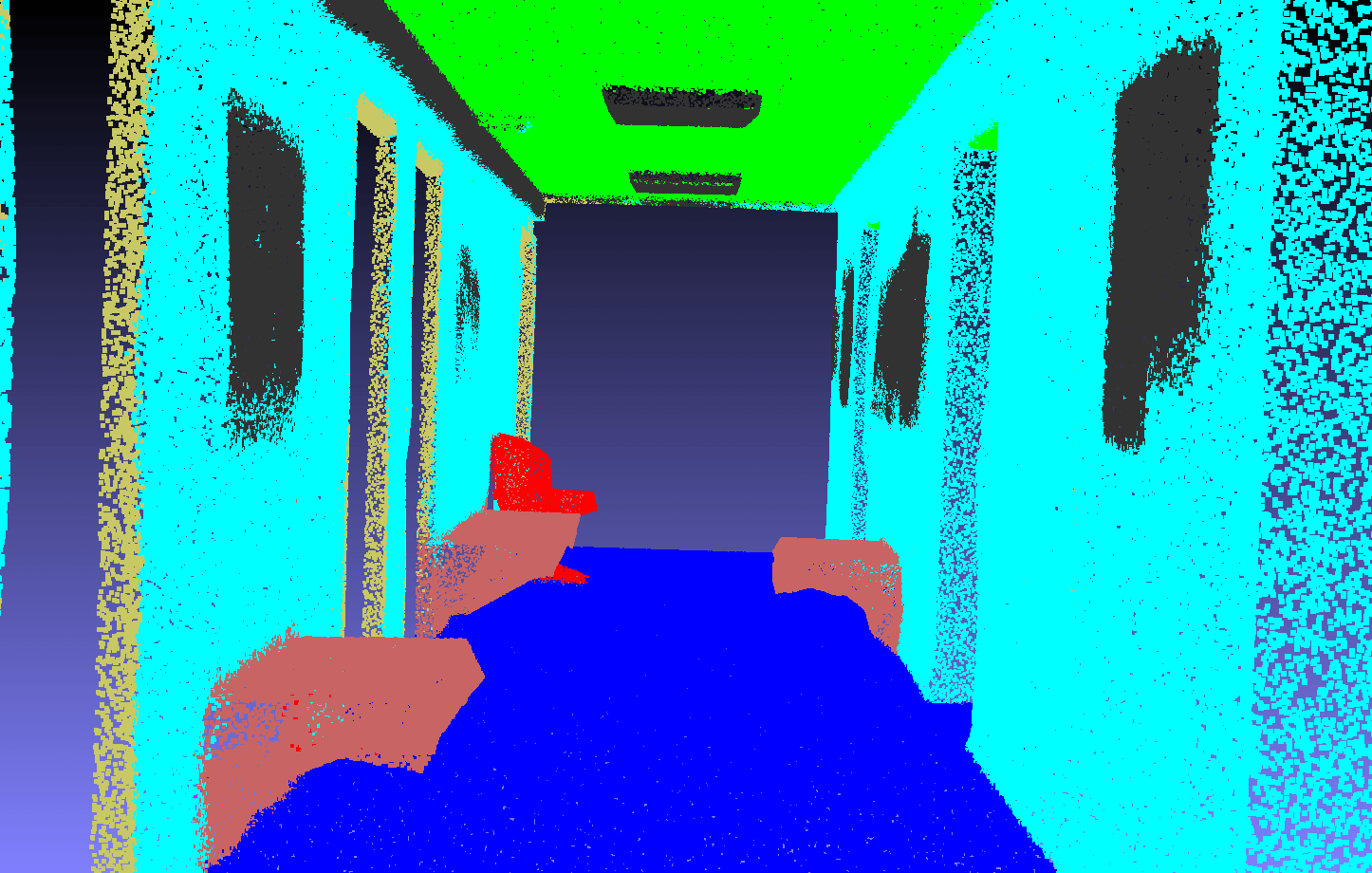}
            \includegraphics[width=2.8cm]{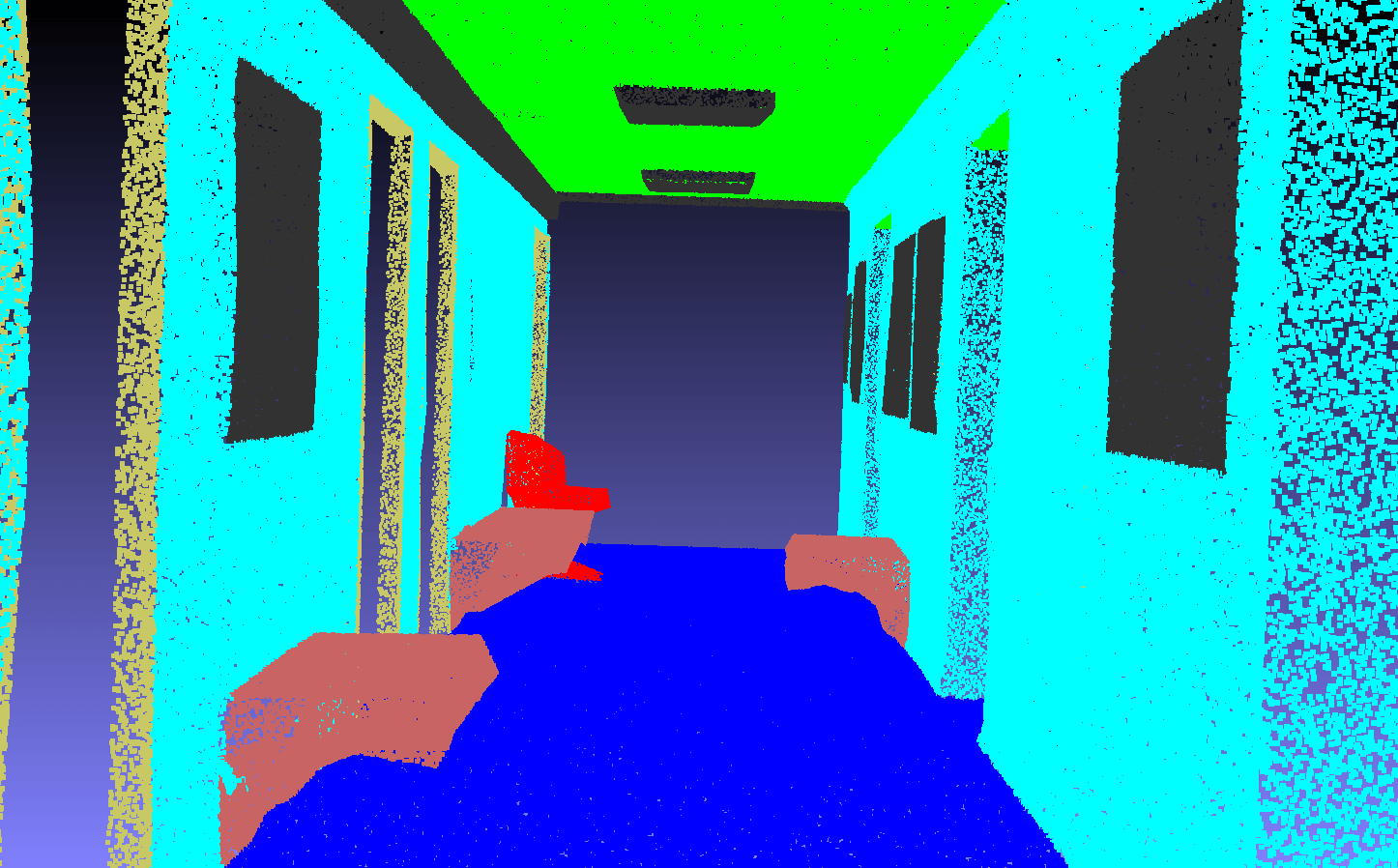}
        
        % % Area5
        %     \includegraphics[width=2.5cm]{Sem_Seg/pointnet2/snapshot04.png}
        %     \includegraphics[width=2.5cm]{Sem_Seg/pointnet2_w_D/snapshot04.png}
        %     \includegraphics[width=2.5cm]{Sem_Seg/gt/snapshot04.png}
        
        % % Area6
        %     \includegraphics[width=2.5cm]{Sem_Seg/pointnet2/snapshot05.png}
        %     \includegraphics[width=2.5cm]{Sem_Seg/pointnet2_w_D/snapshot05.png}
        %     \includegraphics[width=2.5cm]{Sem_Seg/gt/snapshot05.png}
        
        \caption{Semantic Segmentation Result Visualization. From left to right column: PointNet++, PointNet++ (w/ DualNorm), Ground Truth. From top to bottom row: auditorium, lounge, hallway. Zoom in to view the details.}
        \label{fig:Sem_Seg}
    \end{figure}

We evaluate PointNorm on the semantic segmentation task, using S3DIS \cite{armeni20163d} as the benchmark dataset. The S3DIS dataset contains 271 scenes from 6 indoor areas and has 13 semantic labels. In Table \ref{tab:sem_seg}, we compare PointNorm with other methods on S3DIS 6-Fold and S3DIS Area-5. With a 0.037M increase in \#Params, and a 0.05G increase in FLOPs, DualNorm boosts the mIoU, mAcc, and OA for PointNet++ on both S3DIS 6-Fold and Area-5. It is worth mentioning that PointNet++ (w/ DualNorm) has the best OA for S3DIS Area-5, surpassing the recent state-of-the-art PointASML, which has far more \#Params and FLOPs. We also visualize the semantic segmentation results in Fig. \ref{fig:Sem_Seg}. We can see that PointNet++ (w/o DualNorm) is much closer to the ground truth. 
% NOTE: mention with increase of 0.03 parameters (DualNorm), we increase XXX and even achieve the best OA on S3DIS Area-5. 

% e.g., the best score for XXX, surpass/outperform by XXX

% The ScanNet dataset contains indoor point clouds with 21 categories. There are 1513  point clouds for training and 100 for testing. 

% Besides, their landscapes ares less smooth.

% TODO: uncomment this one

\section{Conclusion} \label{sec:conclusion}

In this paper, we propose PointNorm, an point cloud analysis framework that eliminates the need for sophisticated feature extractors. The key ingredient of PointNorm is the DualNorm module, where we address the point cloud irregularity by normalizing the grouped points and the sampled points to each other using local mean and global standard deviation. This straightforward design benefit both the classification accuracy, computational efficiency, loss stability, and gradient stability. Comprehensive experiments and ablation studies demonstrate the effectiveness of the proposed method. In the future, we plan to apply PointNorm to object detection (e.g., SUN RGB-D \cite{song2015sun}) and outdoor semantic segmentation (e.g., SemanticKITTI \cite{behley2019semantickitti}) tasks. It is also interesting to explore semantic information \cite{zheng2022semantic}, mutual information \cite{lu2022unsupervised}, and adversarial similarity \cite{lu2022introvae} in our framework. 

% We sincerely hope this work will inspire the point cloud analysis community to slow down the race for sophisticated feature extractors and complex model architectures and to revisit the succinct design philosophy. We also hope to apply PointNorm to object detection (e.g., SUN RGB-D \cite{song2015sun}) and large-scale outdoor semantic segmentation (e.g., SemanticKITTI \cite{behley2019semantickitti}) tasks. 

% We also hope to apply PointNorm to object detection (e.g., SUN RGB-D \cite{song2015sun}) and semantic segmentation (e.g., S3DIS \cite{armeni20163d}) tasks. 

\bibliographystyle{unsrt}
\bibliography{egbib}

\end{document}